\def\paperID{10905}
\def\confName{ICCV}
\def\confYear{2025}

\def\authorBlock{
    Sijie Zhao\textsuperscript{1,2} \qquad
    Feng Liu\textsuperscript{2,3} \qquad
    Xueliang Zhang\textsuperscript{1*} \qquad
    Hao Chen\textsuperscript{2*} \qquad
    Tao Han\textsuperscript{2} \qquad \\
    Junchao Gong\textsuperscript{2,3} \qquad
    Ran Tao\textsuperscript{2} \qquad
    Pengfeng Xiao\textsuperscript{1} \qquad
    Lei Bai\textsuperscript{2} \qquad
    Wanli Ouyang\textsuperscript{2} \\
    {\small
    \textsuperscript{1}Nanjing University \quad
    \textsuperscript{2}Shanghai Artificial Intelligence Laboratory \quad
    \textsuperscript{3}Shanghai Jiao Tong University} \\
    {\tt\small zsj2233@smail.nju.edu.cn; liufeng2317@sjtu.edu.cn; zxl@nju.edu.cn; chenhao1@pjlab.org.cn}
}

\newif\ifreview 
\newif\ifarxiv \newcommand{\arxiv}{\arxivtrue}
\newif\ifcamera 
\newif\ifrebuttal 

\arxiv 

\pdfoutput=1
\documentclass[10pt,twocolumn,letterpaper]{article}
\ifreview \usepackage[review]{iccv} \fi
\ifarxiv \usepackage[pagenumbers]{iccv} \fi
\ifrebuttal \usepackage[rebuttal]{iccv} \fi
\ifcamera \usepackage{iccv} \fi

\usepackage{graphicx}	
\usepackage{amsmath}	
\usepackage{amssymb}	
\usepackage{booktabs}
\usepackage{times}
\usepackage{microtype}
\usepackage{epsfig}
\usepackage{caption}
\usepackage{float}
\usepackage{placeins}
\usepackage{color, colortbl}
\usepackage{stfloats}
\usepackage{enumitem}
\usepackage{tabularx}
\usepackage{xstring}
\usepackage{multirow}
\usepackage{xspace}
\usepackage{url}
\usepackage{subcaption}
\usepackage{xcolor}
\usepackage[hang,flushmargin]{footmisc}

\ifcamera \usepackage[accsupp]{axessibility} \fi

% ...

\ifarxiv  \fi

\newcommand{\R}[1]{{%
    \textbf{%
        \ifstrequal{#1}{1}{\textcolor{red}{R#1}}{%
        \ifstrequal{#1}{2}{\textcolor{blue}{R#1}}{%
        \ifstrequal{#1}{3}{\textcolor{magenta}{R#1}}{%
        \ifstrequal{#1}{4}{\textcolor{teal}{R#1}}{%
                           \textcolor{cyan}{R#1}%
        }}}}%
    }%
}}

\usepackage{xr-hyper}

\usepackage{multirow}

\makeatletter
\newcommand*{\addFileDependency}[1]{
  \typeout{(#1)}
  \@addtofilelist{#1}
  \IfFileExists{#1}{}{\typeout{No file #1.}}
}

\makeatother
\newcommand*{\myexternaldocument}[1]{
    \externaldocument{#1}
    \addFileDependency{#1.tex}
    \addFileDependency{#1.aux}
}

\definecolor{iccvblue}{rgb}{0.21,0.49,0.74}
\usepackage[pagebackref,breaklinks,colorlinks,allcolors=iccvblue]{hyperref}
\usepackage[capitalize]{cleveref}
\crefname{section}{Sec.}{Secs.}
\crefname{table}{Table}{Tables}
\crefname{figure}{Fig.}{Figs.}

\ifarxiv \crefname{appendix}{App.}{Apps.}
\else \crefname{appendix}{Suppl.}{Suppls.} \fi

\unless\ifarxiv \myexternaldocument{sections/_supplementary} \fi

\begin{document}

\title{Transforming Weather Data from Pixel to Latent Space}

\author{\authorBlock}
\maketitle

\begin{abstract}

    The increasing impact of climate change and extreme weather events has spurred growing interest in deep learning for weather research. However, existing studies often rely on weather data in pixel space, which presents several challenges such as smooth outputs in model outputs, limited applicability to a single pressure-variable subset (PVS), and high data storage and computational costs. To address these challenges, we propose a novel Weather Latent Autoencoder (WLA) that transforms weather data from pixel space to latent space, enabling efficient weather task modeling. By decoupling weather reconstruction from downstream tasks, WLA improves the accuracy and sharpness of weather task model results. The incorporated Pressure-Variable Unified Module transforms multiple PVS into a unified representation, enhancing the adaptability of the model in multiple weather scenarios. Furthermore, weather tasks can be performed in a low-storage latent space of WLA rather than a high-storage pixel space, thus significantly reducing data storage and computational costs. Through extensive experimentation, we demonstrate its superior compression and reconstruction performance, enabling the creation of the ERA5-latent dataset with unified representations of multiple PVS from ERA5 data. The compressed full PVS in the ERA5-latent dataset reduces the original 244.34 TB of data to 0.43 TB. The downstream task further demonstrates that task models can apply to multiple PVS with low data costs in latent space and achieve superior performance compared to models in pixel space. Code, ERA5-latent data, and pre-trained models are available at \href{https://anonymous.4open.science/r/Weather-Latent-Autoencoder-8467}{\textcolor{blue}{URL}}.

\end{abstract}
\section{Introduction}
\label{sec:intro}

The profound impact of climate change and extreme weather events on the Earth has attracted widespread attention \cite{patz_2005_Impact, wild_2025_Climate, chen_2025_Global}. Recently, deep learning methods have made groundbreaking advancements in meteorology, leading to increasing interest in their application to weather research \cite{ravuri_2021_Skilful, liu_2022_Meshless, yang_2023_Predictor, zhang_2023_Skilful, gong2024cascast, gong2024postcast}. However, most existing studies focus primarily on weather-related tasks in the pixel space of weather data \cite{bi_2023_Author, chen_2023_FuXi, chen_2023_FengWu}. Due to the inherent uncertainty of weather-related tasks, the vast diversity of weather data, and the high cost of data storage and processing, prior studies often encounter the dilemma of high data costs and inefficient weather models.

\begin{figure*}[!ht]
        \centering
        \includegraphics[width=\textwidth]{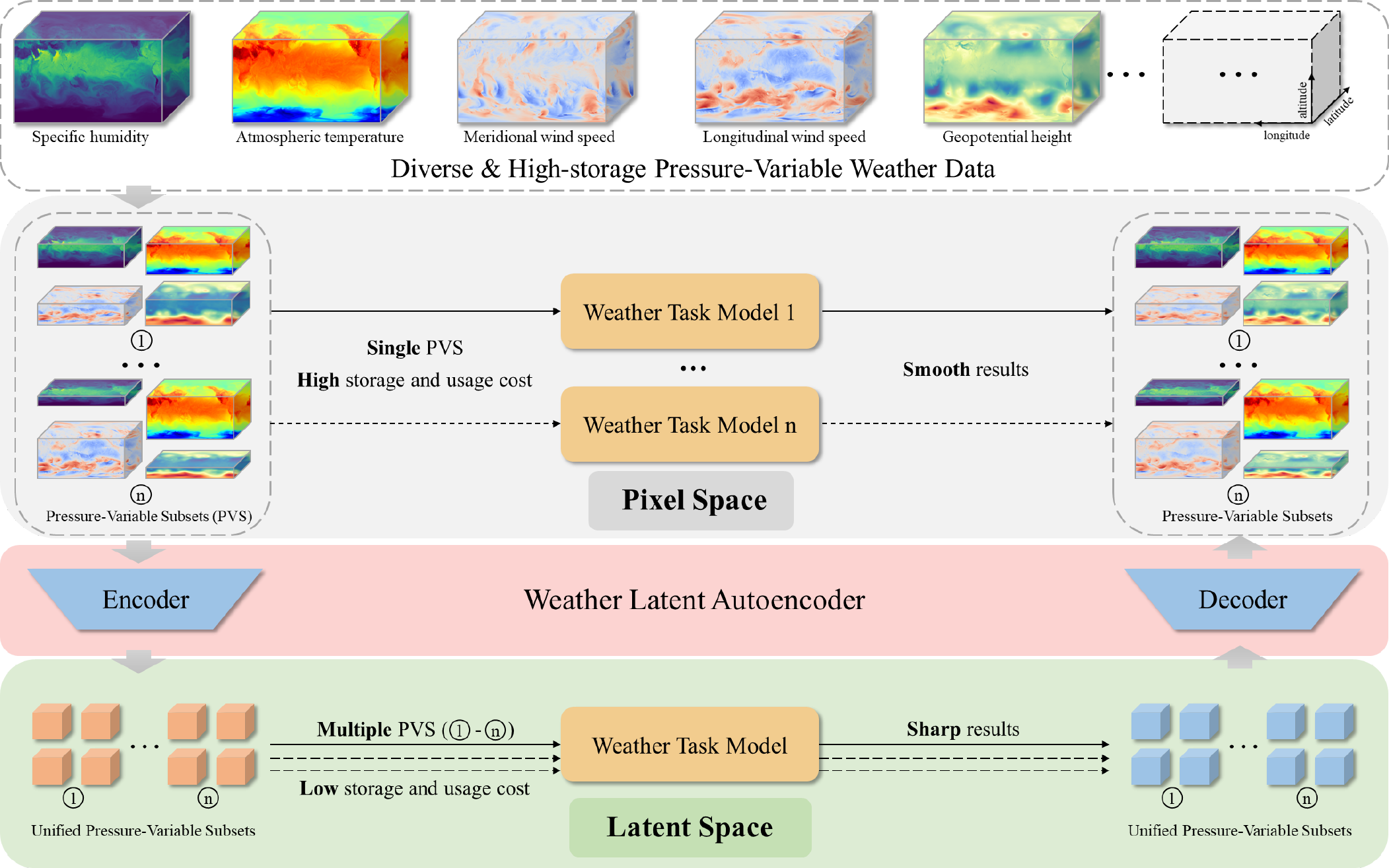}
        \caption{Transforming weather data from diverse and high-storage pixel space to unified and low-storage latent space for weather tasks using weather latent autoencoder. The weather task model in pixel space suffers from high data storage and computational costs and limited applicability to single pressure-variable subset, often yielding ambiguous results. In contrast, the model in latent space benefits from reduced data storage and computational costs, enabling the use of multiple pressure-variable subsets and producing sharper results.}
        \label{fig:overview}
    \end{figure*}

Specifically, performing weather-related tasks in the pixel space of weather data presents three main limitations, as shown in Fig.\ref{fig:overview}: 1) \textbf{Smooth Model Results}. Weather data contain rich small-scale weather structures. When performing tasks such as weather forecasting and downscaling in the pixel space, the model also needs to perform weather reconstruction, requiring a fine reconstruction of small-scale weather structures. However, the inherent uncertainty in weather-related tasks degrades the performance of small-scale weather structure reconstruction and extreme events prediction, leading to smooth results \cite{ravuri_2021_Skilful}. 2) \textbf{Limited Model Applicability to a Single Pressure-Variable Subset (PVS)}. Weather data typically record various weather variables across multiple pressure levels, leading to significant data diversity in the pixel space \cite{astruc_2024_OmniSat, xiong_2024_Neural}. Different weather-related tasks and applications often require distinct PVS selections. For instance, the 500 hPa geopotential height and the 850 hPa wind fields are fundamental in representing atmospheric steering flows and vortex dynamics, which are key to typhoon path prediction \cite{hua_2010_Further, moore_2015_Patterns}. Conversely, the 500 hPa geopotential height, 700 hPa vertical velocity, and 925 hPa specific humidity serve as essential parameters for short-term rainfall forecasting \cite{kuligowski_1998_Experiments, tian_2015_Statistical}. However, models trained in pixel space are typically restricted to a single PVS, limiting their adaptability across multiple weather scenarios requiring different PVS compositions. 3) \textbf{High Data Storage and Computational Costs}. Pixel-based weather datasets can reach hundreds of terabytes (TB) or even petabytes (PB), leading to significant storage and computational costs \cite{hersbach_2020_ERA5}. This poses a substantial challenge for the large-scale application of deep learning in meteorology \cite{klöwer_2021_Compressing,han_2024_CRA5}.

To address the above limitations, we propose a novel approach that transforms weather data from pixel space to latent space for weather-related tasks. Specifically, we introduce the Weather Latent Autoencoder (WLA), a model inspired by VQGAN \cite{esser_2020_Taming}, as illustrated in Fig.\ref{fig:overview}. WLA effectively encodes diverse and high-storage weather data from the pixel space to a unified and lower-storage latent space, facilitating its application to multiple PVSs. This transformation allows weather-task models to operate directly in the latent space, eliminating the need for pixel-space data, thereby enhancing their adaptability to different PVS compositions while significantly reducing data storage and computational costs.

Specifically, WLA addresses the aforementioned issues in three ways: 1) \textbf{Decoupling Weather Reconstruction from Weather Tasks}. In this approach, weather tasks are performed in the latent space, while weather reconstruction occurs within the pretrained Weather Latent Autoencoder. The pretrained WLA ensures that latent features effectively preserve small-scale weather characteristics, allowing for high-quality reconstruction from these features. During weather tasks in the latent space, the uncertainty inherent in these tasks has minimal impact on the small-scale weather structure reconstruction of WLA, resulting in sharp and accurate outcomes for the weather task model. 2) \textbf{Unified Pressure-Variable Representation}. We introduce a Pressure-Variable Unified Module (PVUM) designed to transform any pressure-variable subset to a unified space. PVUM leverages pressure-variable metadata in weather data to generate adaptive mapping network weights through a hypernetwork, enabling the conversion of weather data from pixel space into a unified latent space. This framework allows the weather task model to seamlessly accommodate various types of weather data inputs in the latent space, enhancing its applicability across diverse weather scenarios. 3) \textbf{Latent Space Framework}. We propose the Latent Space Framework, which transitions weather task models from pixel space to latent space, significantly reducing data storage and computational costs. Thanks to WLA's superior compression and reconstruction capabilities, the latent data retain most of the information from the original pixel data, but with a much smaller storage footprint. This results in a substantial reduction in storage costs. Furthermore, tasks such as model training, validation, and testing, which typically require large amounts of data, can be carried out using low-storage latent data, yielding significant savings in data computational costs. 

To promote further research on weather tasks in latent space, we have created a latent dataset based on ERA5 data \cite{hersbach_2020_ERA5}, termed \textbf{ERA5-latent}. Given the large size of ERA5 data, which spans hundreds of TB, many studies opt for Weatherbench as a substitute for ERA5. However, Weatherbench provides only fixed subsets of upper-air variables at 13 pressure levels and 6 surface variables, with a spatial resolution of 1.40505° (128×256 size). This limited resolution and reduced variable scope make it less suitable for contemporary deep learning applications in meteorology. To address these challenges, we employ WLA to transform ERA5 data from pixel space to latent space, thereby creating the ERA5-latent dataset. This transformation not only reduces data costs but also facilitates the study of weather task models across full ERA5 maps and diverse weather data.

The original ERA5 data corresponding to ERA5-latent has a spatial resolution of 0.25° (721×1440 size) and contains 164 weather variables, with a total storage size of 244.34 TB. After transforming these pixel-space data into latent space using WLA, the storage size is reduced to 0.43 TB, achieving a compression ratio of 566.3, thus significantly lowering storage costs. Additionally, to accommodate the diverse data needs of weather task models in multiple scenarios, the ERA5-latent dataset includes unified compressed data for 6 upper-air variables across 6, 13, and 25 pressure levels, as well as surface variables with 4 and 8 variables, and precipitation variables with 1 and 6 variables. Weather task models can leverage the low-storage latent data from ERA5-latent for training, validation, and testing, further minimizing the data computational costs.

In summary, our main contributions are as follows:

\begin{enumerate}
    \item We are the first to propose the novel idea of transforming weather data from pixel space to latent space for weather tasks. By transforming data into latent space, we decouple weather reconstruction from the downstream tasks, enabling the model to generate sharp and accurate results. The unified representation of pressure levels and variables allows task models to handle multiple pressure-variable subsets, while the latent space framework significantly reduces data storage and computational costs.
    \item We introduce the Weather Latent Autoencoder for the pixel-to-latent transformation of weather data. WLA can effectively transform any pressure-variable subset from pixel space to a unified latent space, providing excellent compression and reconstruction performance. This allows weather task models to operate in latent space, achieving high accuracy with low data cost across multiple pressure-variable subsets.
    \item We have constructed the ERA5-latent dataset, which provides large-scale ERA5 weather data with multiple pressure-variable subsets in a smaller data storage footprint and unified latent space. This transformation reduces the data costs of original ERA5 data, offering a robust data foundation for broader meteorological research. 
\end{enumerate}

\section{Related Work}
\label{sec:related}

\subsection*{Managing the Diversity of Weather Data}
    

    Earth science modeling is challenged by heterogeneous observational data. Current methods either rely on specialized architectures such as Omnisat’s modality-specific encoders for cross-modal feature alignment \cite{astruc_2024_OmniSat} or on metadata-driven adaptation, as seen in DOFA’s spectral self-supervision \cite{xiong_2024_Neural}. In weather forecasting, the combinatorial complexity of atmospheric variables and pressure levels often results in brittle models. For instance, FengWu \cite{chen_2023_FengWu} employs 5 upper-air variables at 37 pressure levels with 4 surface variables, while Pangu \cite{bi_2023_Author} and FengWu-GHR \cite{han_2024_FengWuGHR} use 13 pressure levels for similar variables. FuXi \cite{chen_2023_FuXi} uses 5 upper-air variables (13 levels) with an expanded set of 5 surface variables, and Gencast \cite{price_2023_GenCast} scales to 6 variables each. These differences underscore the need for unified frameworks that can flexibly handle diverse pressure-variable subsets.

\subsection*{Weather Data Compression}

    Weather data compression has advanced from traditional linear quantization—exemplified by CAMS’s GRIB2-based 17× compression \cite{inness_2019_CAMS,klöwer_2021_Compressing}—to neural representation learning. Autoencoder-based models \cite{liang_2023_SZ3} and coordinate-aware networks \cite{huang_2022_Compressing} achieve high compression ratios through instance-specific overfitting, though often at the cost of generalization. Meta-learning methods like COIN++ \cite{dupont_2022_COIN} address this by leveraging shared priors for modality-agnostic compression. More recent advances combine probabilistic modeling with entropy coding; for example, Mirowski et al. \cite{mirowski_2024_Neural} achieve 1000× compression using hyperpriors and vector quantization, while CRA5 \cite{han_2024_CRA5} employs a dual-variational transformer to optimize rate-distortion via hierarchical latent space modeling. Our weather latent autoencoder further integrates compression and computation in the latent space, thereby reducing data computational costs.

\subsection*{Low-Cost Weather Datasets}

    The exponential growth of weather data—often reaching petabyte scales—poses significant challenges in storage, computation, and accessibility. Curated low-cost datasets such as Weatherbench \cite{rasp_2020_WeatherBench} mitigate these issues by downsampling ERA5 reanalysis data to a 1.405° resolution (128×256) with 13 pressure levels, cutting storage requirements by 94\% compared to native resolutions. In contrast, CRA5 retains ERA5’s full 0.25° resolution (721×1440) across 159 fields, achieving similar storage efficiency through neural compression at the expense of requiring decoder reconstruction. Collectively, these studies highlight the importance of efficient data representation and unified frameworks for advancing both atmospheric modeling and computer vision applications.

\section{Method}
\label{sec:method}

\begin{figure*}[!ht]
    \centering
    \includegraphics[width=\textwidth]{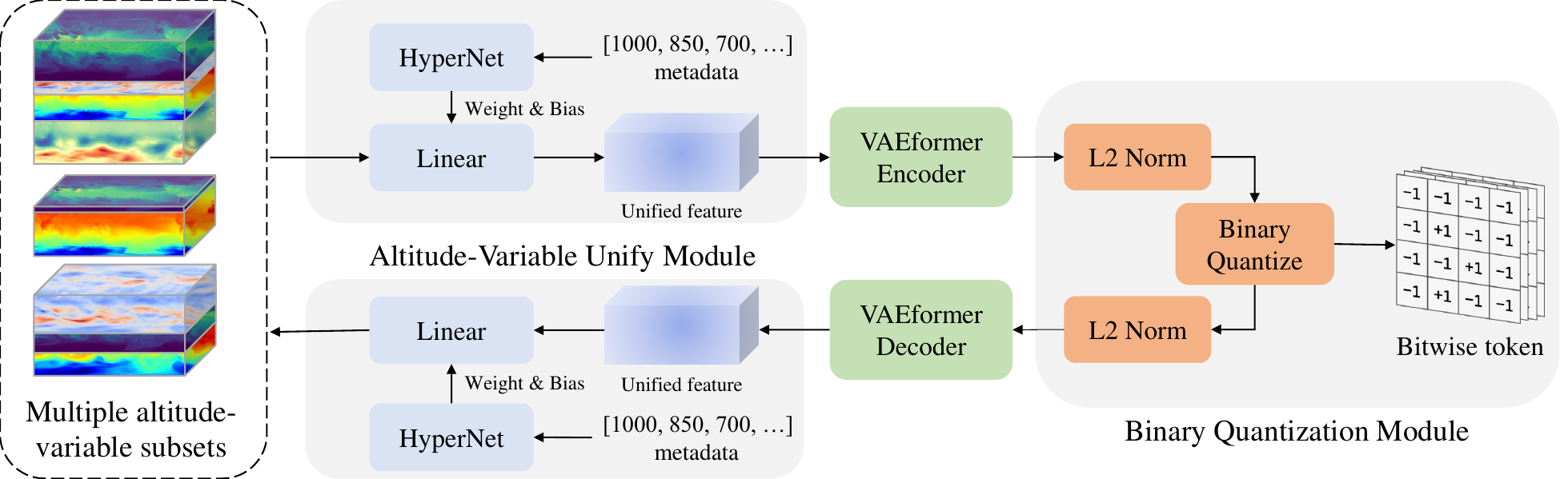}
    \caption{Architecture of the Weather Latent Autoencoder, which compresses weather data from a diverse, high-storage pixel space into a unified, low-storage latent space, and reconstructs it back into the pixel space.}
    \label{fig:WLA_overview}
\end{figure*}

\subsection{Overview of Weather Latent Autoencoder}
\label{sec:overview}
    
    The Weather Latent Autoencoder transforms weather data from diverse and high-storage pixel space into unified and low-storage latent space. As illustrated in Figure~\ref{fig:WLA_overview}, our framework integrates three core components: (1) a Pressure-Variable Unified Module that leverages metadata information to align heterogeneous PVS features, (2) a VAEformer Encoder-Decoder pair adopting the transformer architecture from CRA5's pretraining stage \cite{han_2024_CRA5} for latent feature compression/reconstruction, and (3) a Binary Quantization Module (BQM) that generates compact bitwise tokens through spherical normalization and binary quantization.

    During the compression phase, the model starts with selecting multiple PVSs from a multiple pressure-variables weather dataset. The PVUM first converts pressure-variable metadata into adaptive parameters through a hypernetwork, enabling cross-scale feature alignment across disparate PVS. These unified features are subsequently encoded by the VAEformer encoder into low-dimensional latent representations, preserving essential weather patterns while discarding pixel-space redundancies. The BQM then projects the latent features onto a unit spherical space through L2-normalization and applies binary quantization to produce storage-efficient bitwise tokens. This compression effectively reduces data storage compared to original pixel-space PVS representations.

    The reconstruction phase executes an inverse transformation through three cascaded operations. Initially, bitwise tokens are mapped back to spherical space via L2-normalization. The VAEformer Decoder subsequently reconstructs unified features through upsampling operators, ensuring high-quality weather data reconstruction. Finally, the PVUM regenerates the original PVS by applying metadata-guided inverse transformations, thereby completing the latent-to-pixel space transformation cycle.

    The latent space framework of WLA offers three fundamental benefits. First, the unified encoding enables weather-task models to directly operate on a unified latent space, eliminating structural modifications for cross-PVS generalization. Second, the WLA decouples weather reconstruction from task modeling. The uncertainty present in the weather tasks does not affect the weather reconstruction, ensuring that the task model can output sharp and accurate results. Third, data storage and computational costs are significantly reduced as model training, validation, and inference primarily utilize low-storage latent features, restricting pixel-space operations to final metric evaluation phases. This layered approach addresses critical challenges in weather data processing— including multiple pressure-variable representation learning, storage scalability, and task-specific adaptation— through systematic latent-space engineering.

\subsection{Pressure-Variable Unified Module}
\label{sec:PVUM}
    
    To map any pressure-variable subset from pixel space to a unified feature, we designed the Pressure-variable Unified Module, which utilizes the metadata of the pressure-variable subset to generate adaptive weights and biases for a linear layer, thereby enabling adaptive feature mapping. 
    
\begin{figure}[!ht]
        \centering
        \includegraphics[width=0.48\textwidth]{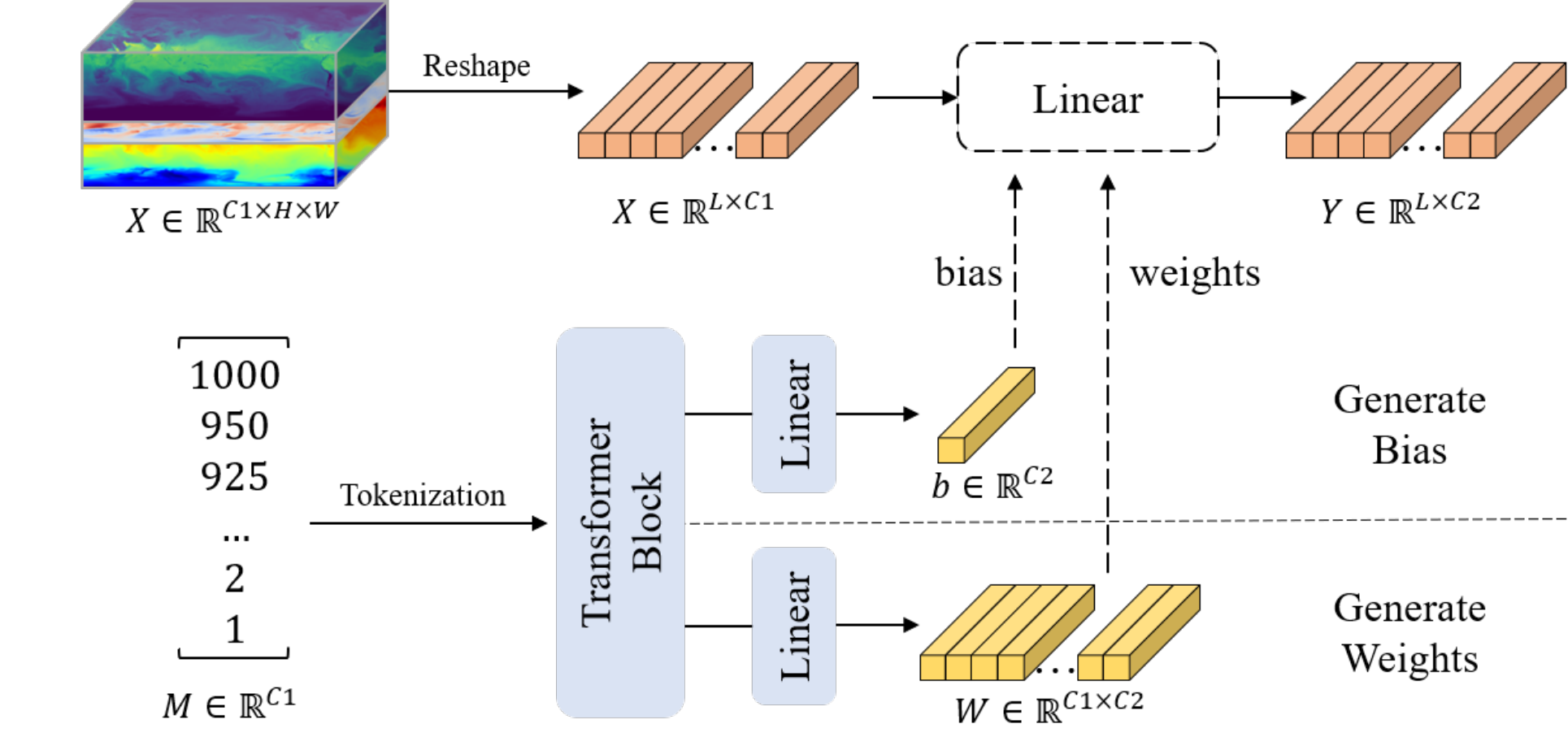}
        \caption{Workflow of Pressure-Variable Unified Module, which transforms diverse weather data into unified representation.}
        \label{fig:PVUM}
    \end{figure}

    As shown in Figure~\ref{fig:PVUM}, given an input PVS tensor \( X \in \mathbb{R}^{C_1 \times H \times W} \) with its pressure-variable metadata \( M \in \mathbb{R}^{C_1} \) (where \( C_1 \) varies across tasks and scenarios), PVUM generates a unified feature \( Y \) with fixed dimensionality through hypernetwork-based parameter generation. This process consists of three core operations:
        \noindent\textbf{Metadata Embedding:} The variable metadata \( M \) containing physical attributes (pressure levels and variables) undergoes positional encoding followed by tokenization. A learnable class token [CLS] is prepended to the token sequence \(T \in \mathbb{R}^{(C_1 + 1) \times d}\), where \( d\) is the embedding dimension.
        \noindent\textbf{Cross-Variable Relation Modeling:} The token sequence passes through several transformer blocks for learning the relationships between the metadata.
        \noindent\textbf{Adaptive Parameter Generation:} The [CLS] token produces bias parameters \( b \in \mathbb{R}^{C_2} \) via a linear projection, while the remaining tokens generate a weight matrix \( W \in \mathbb{R}^{C_1 \times C_2} \) through another linear layer. The resulting weights \( W \) and bias \( b \) form a linear layer that maps the features with \( C_1 \) channels to features with \( C_2 \) channels. Therefore, the input \( X \) is reshaped from \( (C_1, H, W) \) to \( (L, C_1) \), where \( L = H \times W \), and then mapped to the target feature \( Y \) with shape \( (L, C_2) \) using the generated linear layer.

    The PVUM structure not only adaptively maps any pressure-variable subset to a unified feature in terms of shape, but also effectively preserves the relationships between weather data in pixel space. Due to the continuity, smoothness, and vertical mixing of the atmosphere, there is inherent similarity between different weather pressure levels and variable data, especially between adjacent pressure levels for the same variable \cite{zhang2023increased}. The hypernetwork of PVUM learns this relationship when modeling the metadata, allowing it to map similar weather variables and adjacent pressure levels to similar unified features. As a result, PVUM preserves the relationships between weather data in the feature space.

\subsection{Binary Quantization Module}
\label{sec:BQM}

    To effectively compress weather features while preserving critical information, we propose the Binary Quantization Module that establishes a bi-directional mapping between continuous features and discrete binary tokens. As shown in Figure~\ref{fig:WLA_overview} (right), the module inherits the vector quantization framework from BSQ \cite{zhao_2024_Image} which has two key components: (1) spherical space projection for stable entropy loss estimation, and (2) deterministic binary quantization for hardware-friendly storage. The quantization process consists of three stages: First, input features undergo L2 normalization to project them onto a spherical space, which not only stabilizes the subsequent quantization but also enables computation of the entropy loss with acceptable memory/space cost \cite{han2024infinity}. Second, we apply sign-based binary quantization where positive values are mapped to 1 and negative values to -1, generating compact bitwise tokens. During reconstruction, the bitwise tokens are inversely projected to the spherical space through L2 normalization before being fed to the VAEformer decoder for upsampling.

    The compression ratio of our weather latent autoencoder can be formally analyzed through the data storage. Let the input feature tensor \( F \in \mathbb{R}^{C \times H \times W}\) with float32 representation be compressed into binary tokens \( B \in \{ -1, 1\}^{C' \times H' \times W'}\). The spatial downsampling factors \( (P_h, P_w) = (H/H', W/W')\) combined with channel dimension adjustment yield a compression ratio
    \begin{align}
    \label{func:comp_ratio}
        R = \frac{C \cdot H \cdot W \cdot 32}{C' \cdot H' \cdot W'} = \frac{C}{C'} \cdot 32 \cdot P_h \cdot P_w.
    \end{align}
    
    
    

\subsection{Latent Space Framework}
\label{sec:LSF}

\begin{figure}[!ht]
        \centering
        \includegraphics[width=0.48\textwidth]{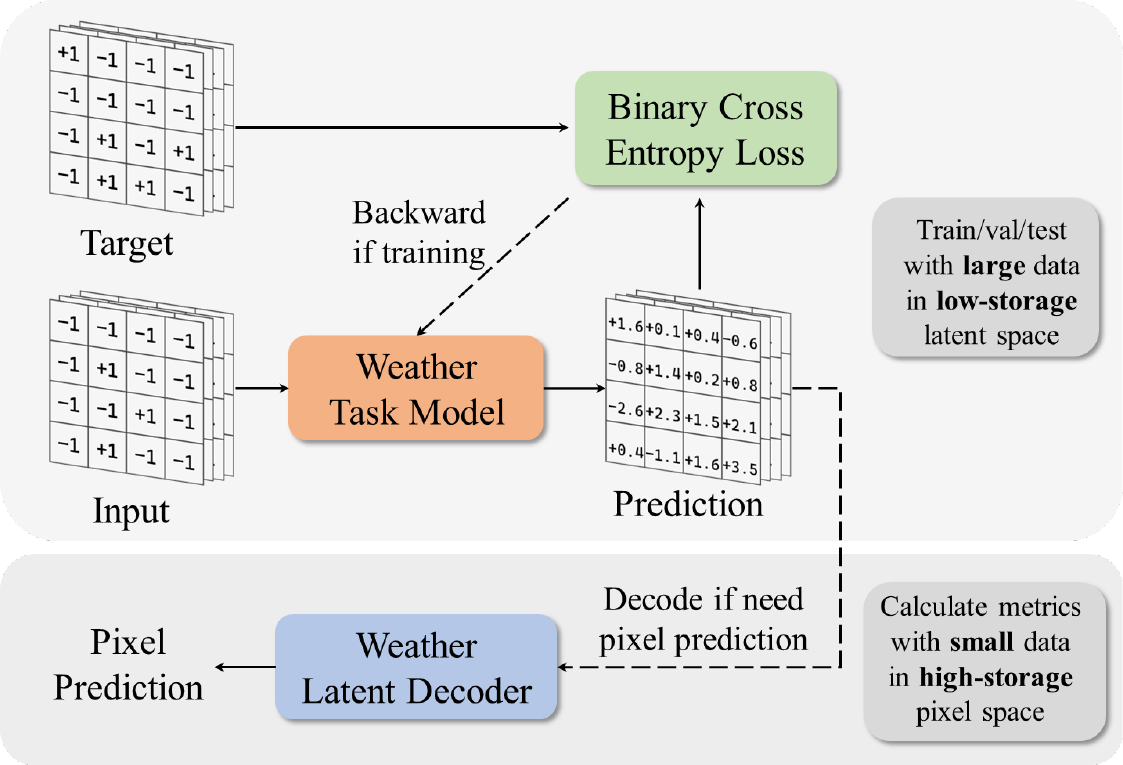}
        \caption{Overview of the Latent Space Framework. The data-intensive processes can be performed in the low-storage latent space, while processes requiring a smaller amount of data can be carried out in the high-storage pixel space, thereby effectively reducing data costs.}
        \label{fig:LSF}
    \end{figure}

    To reduce the data storage and computational costs when using deep learning models for weather tasks, we propose the Latent Space Framework as illustrated in Figure~\ref{fig:LSF}. Given that the Weather Latent Autoencoder has effective compression and reconstruction of weather data, our framework leverages two key observations: (1) Latent representations preserve essential information from pixel-space data, and (2) Data similarity relationships remain consistent across both pixel and latent spaces. Consequently, processes that require large amounts of data, such as training, validation, and testing of weather models, can be conducted in the lower-storage latent space.

    Specifically, when conducting weather tasks, the input bitwise token passes through the weather task model, and the output prediction is compared to the target to compute the binary cross-entropy loss. During the training phase, the loss is used for gradient backpropagation and parameter updates. During the validation and testing phases, model performance can be directly evaluated on the low-storage latent space. In processes that only contain a small amount of data, such as calculating pixel metrics for weather tasks, we use the weather latent decoder to decode and reconstruct the predictions back to the high-storage pixel data, from which the corresponding pixel metrics can be calculated.

    By utilizing the Latent Space Framework, we can store large amounts of data in the low-storage latent space and small amounts of data in the high-storage pixel space, effectively reducing data storage costs. Additionally, weather task models can be trained, validated, and tested in the latent space, where large amounts of data are required, while pixel space can be used for processes like pixel metrics calculation that need only a small amount of data, thereby lowering the data computational costs.

\section{Experimental Results}
\label{sec:experimental results}
    
\subsection{Dataset}
\label{sec:datasets}

    The ERA5 dataset \cite{hersbach_2020_ERA5} is a global atmospheric reanalysis product from the European Center for Medium-Range Weather Forecasts (ECMWF) and serves as a standard for evaluating the Weather Latent Autoencoder in comparison to other models. This dataset is highly valued in climate research due to its high spatial resolution of 0.25° and extensive weather coverage. To meet various weather-related needs, we organized three categories of variables: upper-air, surface, and precipitation variables. For \textbf{upper-air variables}, we selected three configurations spanning 25, 13, and 6 pressure levels, each containing six core weather variables: geopotential height ($z$), longitudinal wind speed ($u$), meridional wind speed ($v$), vertical velocity ($w$), atmospheric temperature ($t$), and specific humidity ($q$), in which variables are presented by abbreviating their short name and pressure levels (e.g., q1000 denotes the specific humidity at a pressure level of 1000 hPa). The \textbf{surface variables} include two subsets: an 8-variable set comprising 10m v-component of wind (10v), 10m u-component of wind (10u), 100m v-component of wind (100v), 100m u-component of wind (100u), 2m temperature (t2m), Total cloud cover (tcc), surface pressure (sp) and Mean sea-level pressure (msl); and a streamlined 4-variable subset (10v, 10u, tcc, msl). The \textbf{precipitation variables} cover cumulative hourly precipitation over six intervals (tp1h, tp2h, tp3h, tp4h, tp5h, tp6h), with two additional single-variable subsets (tp1h and tp6h).  
    
    The dataset is temporally partitioned into training sets (1979–2021, 233.48 TB), validation sets (2022, 5.43 TB), and test sets (2023, 5.43 TB). To optimize domain-specific performance, we designed distinct WLA architectures: individual WLAs are trained for each upper-air variable to capture multi-pressure-level dependencies, while unified WLAs handle surface and precipitation variables to exploit intra-category correlations. This hierarchical strategy ensures balanced reconstruction fidelity and computational efficiency across heterogeneous weather data structures.

\subsection{Implementation Details}
    
    All WLAs are trained with identical configurations across weather variables. The models are optimized for 500K steps on 4 Tesla A100 GPUs. The codebook size is set to \(2^{128}\) for upper-air and surface variables, while precipitation variables employ a reduced codebook size of \(2^{32}\) due to their higher compressibility. The input data are processed through patches of size \(15 \times 14\) with a stride of \(10 \times 10\) and a padding of \(2 \times 2\).  
    
    Following the architectural insights of \cite{hansen-estruch_2025_Learnings}, where decoder upscaling demonstrated significant reconstruction benefits without comparable encoder gains, we design the VAEformer with asymmetric depths: a 16-layer encoder versus a 32-layer decoder. Training employs the AdamW \cite{loshchilov2017decoupled} optimizer with an initial learning rate of \(3.2 \times 10^{-5}\), batch size 8, and a hybrid learning schedule that combines a linear warm-up phase increasing the learning rate from \(3.2 \times 10^{-6}\) to \(3.2 \times 10^{-5}\), followed by a cosine decay phase.  

\begin{table*}[!ht]
\caption{
    Compression Result of WLA and several state-of-the-art compression methods.
}
\begin{tabular}{lcclllccccc}
\hline
\multirow{3}{*}{Method}           & \multicolumn{8}{c}{Weighted RMSE $\downarrow$}                                                                                                                                         & \multirow{3}{*}{\begin{tabular}[c]{@{}c@{}}Comp. \\ Ratio $\uparrow$\end{tabular}} & \multirow{3}{*}{bpsp $\downarrow$} \\ \cline{2-9}
                                  & \multicolumn{5}{c}{Upper-air Variables}                                                                    & \multicolumn{2}{l}{Surface Variables} & \multicolumn{1}{l}{Precipitation} &                                                                                    &                                    \\
                                  & w500           & \multicolumn{2}{c}{w700}           & \multicolumn{1}{c}{q700} & \multicolumn{1}{c}{q1000} & TCC               & SP                & tp6h                              &                                                                                    &                                    \\ \hline
Var. Std (ref.)                   & 0.218          & \multicolumn{2}{c}{0.240}          & 0.0025                   & 0.0059                    & 0.36              & 9584.49           & 1.57                              & --                                                                                 & --                                 \\ \hline
Elic \cite{he_2022_ELIC}          & 0.197          & \multicolumn{2}{c}{0.233}          & 0.00076                  & 0.00087                   & 0.18              & 537.82            & 1.19                              & 648.3                                                                              & 0.112                              \\
IEN \cite{xie_2021_Enhanced}      & 0.213          & \multicolumn{2}{c}{0.247}          & 0.00084                  & 0.00092                   & 0.23              & 688.27            & 1.03                              & 202.5                                                                              & 0.158                              \\
VQVAE \cite{mirowski_2024_Neural} & 0.382          & \multicolumn{2}{c}{0.401}          & 0.00108                  & 0.00113                   & 0.19              & 673.32            & 1.29                              & \textbf{1100.0}                                                                    & \textbf{0.029}                     \\
VQGAN \cite{mirowski_2024_Neural} & 0.367          & \multicolumn{2}{c}{0.371}          & 0.00101                  & 0.00107                   & 0.18              & 652.38            & 1.20                              & \textbf{1100.0}                                                                    & \textbf{0.029}                     \\
VAEformer \cite{han_2024_CRA5}    & 0.117          & \multicolumn{2}{c}{0.134}          & 0.00031                  & 0.00035                   & 0.12              & 376.90            & 0.80                              & 323.1                                                                              & 0.099                              \\
\textbf{WLA (Upper-air)}          & \textbf{0.076} & \multicolumn{2}{c}{\textbf{0.083}} & \textbf{0.00027}         & \textbf{0.00028}          & --                & --                & --                                & 625.9                                                                              & 0.051                              \\
\textbf{WLA (Surface)}            & --             & \multicolumn{2}{c}{--}             & \multicolumn{1}{c}{--}   & \multicolumn{1}{c}{--}    & \textbf{0.055}    & \textbf{257.88}   & --                                & 200.3                                                                              & 0.159                              \\
\textbf{WLA (Precipitation)}      & \textbf{--}    & \multicolumn{2}{c}{--}             & \multicolumn{1}{c}{--}   & \multicolumn{1}{c}{--}    & --                & --                & \textbf{0.47}                     & 600.9                                                                              & 0.053                              \\ \hline
\end{tabular}
\label{tab:compression_results}
\end{table*}

\subsection{Ablation Study}

    The Weather Latent Autoencoder enables the mapping of multiple pressure-variable subsets into a unified latent space, where the compression ratio and reconstruction performance are influenced by the number of input variables and the codebook size. To investigate and quantify their impacts for identifying an optimal balance between compression efficiency and reconstruction quality, we conducted ablation experiments on the atmospheric temperature variable of the upper-air dataset. Specifically, we evaluated three configurations of input pressure levels (6, 13, and 25 layers) and five codebook sizes (\(2^{16}, 2^{32}, 2^{64}, 2^{96}, 2^{128}\)).

    The ablation results, illustrated in Figure 5, reveal that the compression ratio inversely correlates with codebook size and positively correlates with the number of input pressure levels, consistent with Equation~\ref{func:comp_ratio}. Conversely, reconstruction quality exhibits a positive correlation with codebook size and a negative correlation with input pressure levels. This indicates that reconstruction performance is proportional to the compressed feature dimensionality but inversely proportional to the input data size, thereby establishing an inherent trade-off between reconstruction quality and compression ratio. Since input data specifications are dictated by weather tasks and scenarios, adjusting the codebook size serves as the primary mechanism to balance this trade-off.

    In this work, we selected a codebook size of \(2^{128}\). Increasing the codebook size beyond this value yields only marginal improvements in reconstruction quality. At this configuration, the WLA achieves compression ratios of 150.2, 325.4, and 625.9 for 6, 13, and 25 pressure levels, respectively, reducing the training dataset to 0.4 TB. This configuration effectively minimizes storage and computational costs while maintaining high reconstruction fidelity. Further expansion of the codebook size would degrade the compression ratio, substantially increasing data storage requirements. Thus, the codebook size of \(2^{128}\) represents an optimal equilibrium between compression efficiency and reconstruction performance on the ERA5 dataset.

    \begin{figure}[!ht]
        \centering
        \includegraphics[width=0.48\textwidth]{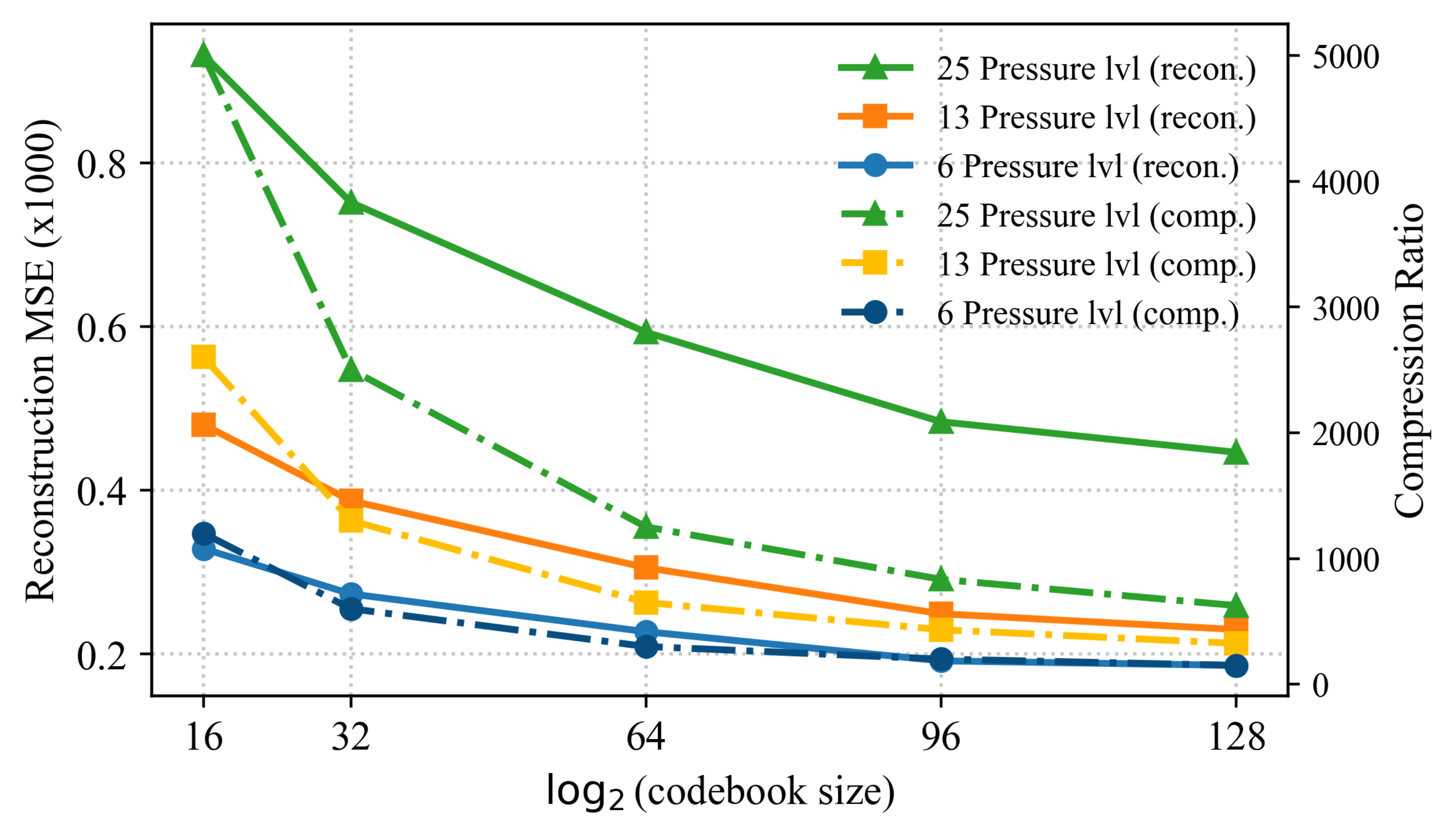}
        \caption{Ablation study on compression ratio and reconstruction quality of the WLA under varying input pressure levels (6, 13, 25 layers) and codebook sizes (\(2^{16}\) to \(2^{128}\)), evaluated on the atmospheric temperature variable.}
        \label{fig:ablation}
    \end{figure}

\subsection{Results}
        
    To demonstrate the effectiveness of the WLA in data compression, we compared it with several state-of-the-art compression methods (Elic \cite{he_2022_ELIC}, IEN \cite{xie_2021_Enhanced}, VQVAE \cite{van2017neural}, VAGAN \cite{esser_2020_Taming}, VAEformer \cite{han_2024_CRA5}) across three metrics: compression ratio, bits per sub-pixel (bpsp) \cite{mentzer_2019_Practical}, and weighted RMSE \cite{han_2024_FengWuGHR} on representative upper-air variables, surface variables, and precipitation variables. The results are summarized in Table~\ref{tab:compression_results}. Due to significant differences in numerical ranges among weather variables, we included a "Variable Std" row as a reference. Generally, variables with higher variances exhibit larger reconstruction errors. Since the number of input variables in WLA influences both the compression ratio and reconstruction quality, we evaluated its performance using the maximum input configurations: 25 pressure levels for upper-air variables, 8 variables for surface variables, and 6 variables for precipitation variables. The compression ratio and bpsp values reported for WLA in Table~\ref{tab:compression_results} correspond to its performance on these three variable categories, respectively. 
    
    As shown in Table~\ref{tab:compression_results}, WLA achieves superior overall compression performance across upper-air, surface, and precipitation variables compared to existing methods, characterized by higher compression ratios, lower bpsp values, and competitive weighted RMSE scores. This demonstrates that WLA effectively balances efficient weather data compression with high-fidelity reconstruction. Notably, WLA not only excels in handling specific variables but also demonstrates remarkable flexibility and versatility, enabling seamless adaptation to diverse variable combinations and complex application scenarios. The visualization results of WLA can be found in Section A of the supplementary materials.


\section{ERA5-Latent Dataset}
\label{sec:latent_datasets}

\subsection{Overview of ERA5-Latent Dataset}

Leveraging the excellent compression and reconstruction performance of the Weather Latent Autoencoder, we transformed the multiple PVSs of ERA5 data described in Section \ref{sec:datasets} from pixel space to latent space, yielding the ERA5-latent dataset. The partitioning of data and the selection of PVS remain consistent with Section \ref{sec:datasets}. By utilizing the high compression rate of WLA, the ERA5-latent dataset reduces the original 244.34 TB of data down to 0.43 TB, while providing a unified representation for multiple PVS. These subsets include three for upper-air variables corresponding to \(25\), \(13\), and \(6\) pressure levels, two for surface variables (4 and 8 variables), and three for precipitation variables (\(\text{tp1h}\), \(\text{tp6h}\), \(\text{tp1-6h}\)). To facilitate the computation of pixel metrics, the ERA5-latent dataset also incorporates the raw pixel data for July 2023, which has been compressed using the Lempel-Ziv-Markov chain-Algorithm (LZMA) and occupies 0.117 TB of storage. LZMA is a widely used lossless compression algorithm developed by Igor Pavlov and implemented via the Python standard library.

Building upon the ERA5-latent dataset, deep learning models for large-scale meteorological research can utilize the unified latent representation to seamlessly handle multiple PVS, making the models adaptable for a wide range of meteorological tasks and scenarios. Moreover, processes that typically require large datasets, such as training, validation, and testing, can be conducted using the compact latent data, significantly reducing both storage and data computational costs. For tasks requiring only a limited amount of pixel data, pixel data from a single month within the ERA5-latent dataset can be used to compute metrics and visually compare model outputs with the original data.

\subsection{Downstream: Weather Forecasting}

To demonstrate that models operating in a unified latent space can adapt to multiple PVS and generate sharper results compared to pixel-space models, we conducted experiments on a weather forecasting task using both the pixel-space model (Window-ViT, WT) and its latent-space variant (Window-ViT-Latent, WTL). The experiments were performed on the ERA5 dataset and the ERA5-latent dataset, respectively. Specifically, following the settings in \cite{bi_2023_Author}, we conduct weather forecasting at 6‐hour intervals. For the Window-ViT model, we use 13 pressure levels for five upper-air variables along with four surface variables. In contrast, to evaluate the Window-ViT-latent's adaptability to multiple PVS, we experiment with multiple pressure levels (25, 13, 6) for the upper-air variables for the latent model. The Window-ViT model employs a Vision Transformer (ViT) as its backbone, utilizes window-based self-attention, and adopts the multimodal encoder-decoder configuration described in \cite{chen_2023_FengWu}. In contrast, the Window-ViT-latent model omits the downsampling and upsampling components present in Window-ViT while keeping all other components unchanged.

\begin{figure}[!ht]
    \centering
    \includegraphics[width=0.48\textwidth]{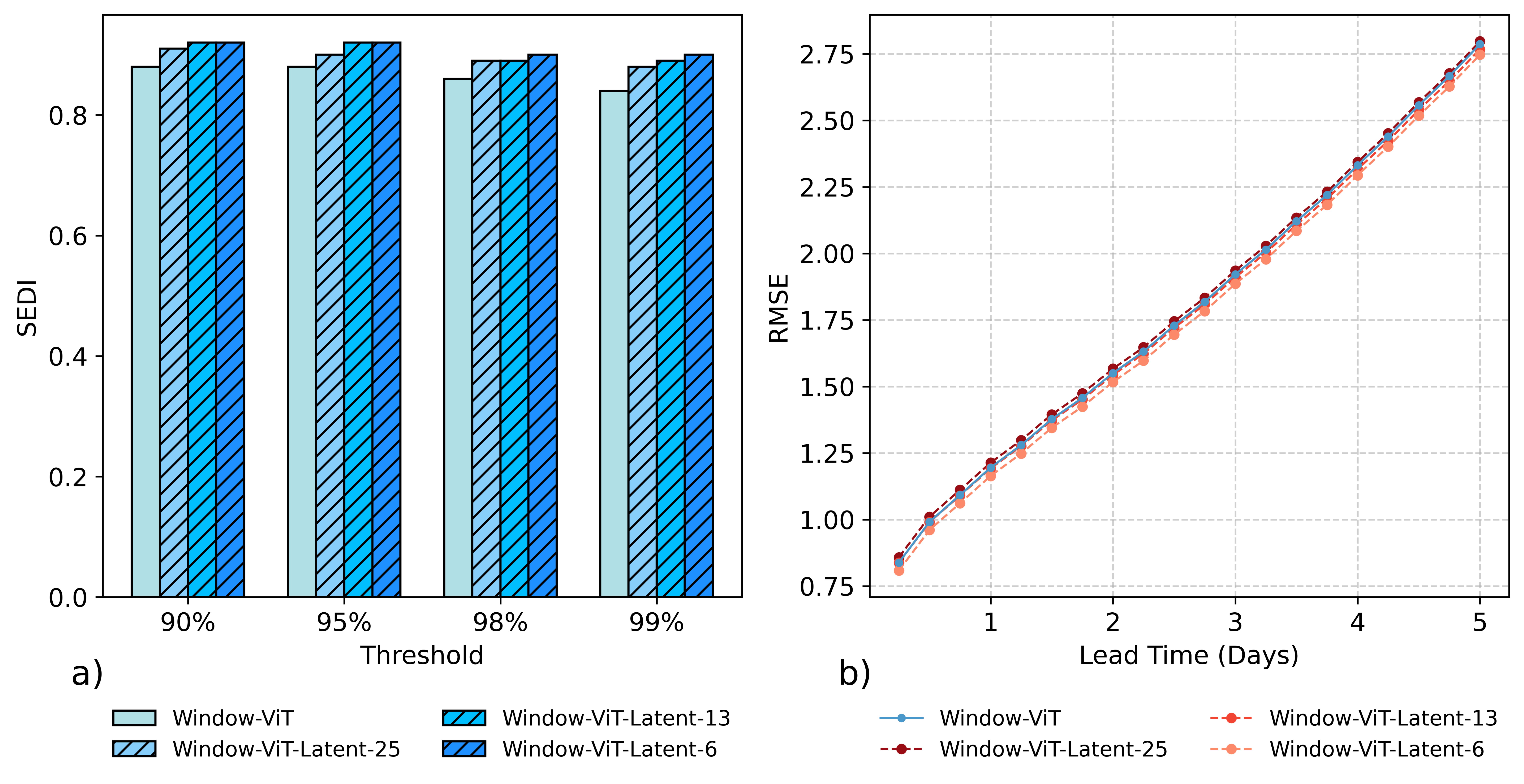}
    \caption{Comparison of weather forecasting performance using the Window-ViT and Window-ViT-latent models on the ERA5 and ERA5-latent datasets. (a) Symmetric Extremal Dependency Index (SEDI) for extreme weather predictions. (b) Root Mean Square Error (RMSE) for overall weather forecasting over a 5-day period. The number following "Window-ViT-latent" indicates the number of pressure levels used in the model.}
    \label{fig:ERA5_latent}
\end{figure}

Using the representative T850 variable as an example, we computed the forecast model’s Symmetric Extremal Dependency Index (SEDI) \cite{kurth2023fourcastnet, xu2024extremecast} at the 6th hour and the RMSE over the subsequent five days, as shown in Fig.\ref{fig:ERA5_latent}. SEDI classifies each pixel into extreme or normal weather using high quantile thresholds (90\%, 95\%, 98\%, and 99\%) and then calculates the hit rate, a value closer to 1 indicates a more accurate prediction of extreme weather.

The results in Fig.\ref{fig:ERA5_latent} (a) demonstrate that Window-ViT-latent yields more accurate extreme weather forecastings and produces sharper outcomes compared to Window-ViT, while the results in Fig.\ref{fig:ERA5_latent} (b) indicate that Window-ViT-latent performs comparably to Window-ViT in overall weather forecasting. Moreover, Window-ViT-latent can accommodate multiple PVS inputs, generating forecasts that are both sharp and accurate. Therefore, employing deep learning models in the latent space for weather tasks can achieve performance on par with pixel-space models while reducing data storage and computational costs, and effectively preserving extreme values in the forecasts.

\section{Conclusion}
\label{sec:conclusion}

In this work, we introduce a novel approach that transforms weather data from pixel space to latent space for weather tasks. This transformation addresses several challenges inherent in pixel space models, including smooth and inaccurate model outputs, limited applicability to a single PVS, and high data storage and computational costs. To overcome these issues, we propose the Weather Latent Autoencoder. By decoupling weather tasks from reconstruction, mapping multiple PVS into a unified latent space, and transferring data-intensive processes into a compact latent space, the Weather Latent Autoencoder can make weather task models generate sharp and accurate results, adapt to any PVS, and significantly reduce storage and computational costs.

Leveraging Weather Latent Autoencoder, we transformed the ERA5 data from pixel space into latent space to construct the ERA5-latent dataset. This dataset provides a unified representation of multiple PVS and can compress 244.34 TB of raw data into 0.43 TB. We believe that both the Weather Latent Autoencoder and the ERA5-latent dataset will serve as robust foundations for extensive future meteorological research in latent space. In future work, we plan to explore the development of autoencoders with improved reconstruction performance and to apply our framework to weather data with higher spatial resolution.


{\small
\bibliographystyle{ieeenat_fullname}
\bibliography{sections/references}

\begin{thebibliography}{41}
\providecommand{\natexlab}[1]{#1}
\providecommand{\url}[1]{\texttt{#1}}
\expandafter\ifx\csname urlstyle\endcsname\relax
  \providecommand{\doi}[1]{doi: #1}\else
  \providecommand{\doi}{doi: \begingroup \urlstyle{rm}\Url}\fi

\bibitem[Astruc et~al.(2024)Astruc, Gonthier, Mallet, and Landrieu]{astruc_2024_OmniSat}
Guillaume Astruc, Nicolas Gonthier, Clement Mallet, and Loic Landrieu.
\newblock Omnisat: Self-supervised modality fusion for earth observation, 2024.

\bibitem[Bi et~al.(2023)Bi, Xie, Zhang, Chen, Gu, and Tian]{bi_2023_Author}
Kaifeng Bi, Lingxi Xie, Hengheng Zhang, Xin Chen, Xiaotao Gu, and Qi Tian.
\newblock Author correction: Accurate medium-range global weather forecasting with 3d neural networks.
\newblock \emph{Nature}, 621\penalty0 (7980):\penalty0 E45--E45, 2023.

\bibitem[Chen et~al.(2023{\natexlab{a}})Chen, Han, Gong, Bai, Ling, Luo, Chen, Ma, Zhang, Su, Ci, Li, Yang, and Ouyang]{chen_2023_FengWu}
Kang Chen, Tao Han, Junchao Gong, Lei Bai, Fenghua Ling, Jing-Jia Luo, Xi Chen, Leiming Ma, Tianning Zhang, Rui Su, Yuanzheng Ci, Bin Li, Xiaokang Yang, and Wanli Ouyang.
\newblock Fengwu: Pushing the skillful global medium-range weather forecast beyond 10 days lead, 2023{\natexlab{a}}.

\bibitem[Chen et~al.(2023{\natexlab{b}})Chen, Zhong, Zhang, Cheng, Xu, Qi, and Li]{chen_2023_FuXi}
Lei Chen, Xiaohui Zhong, Feng Zhang, Yuan Cheng, Yinghui Xu, Yuan Qi, and Hao Li.
\newblock Fuxi: A cascade machine learning forecasting system for 15-day global weather forecast.
\newblock \emph{npj Climate and Atmospheric Science}, 6\penalty0 (1):\penalty0 190, 2023{\natexlab{b}}.

\bibitem[Chen et~al.(2025)Chen, Brun, Buri, Fatichi, Gessler, McCarthy, Pellicciotti, Stocker, and Karger]{chen_2025_Global}
Liangzhi Chen, Philipp Brun, Pascal Buri, Simone Fatichi, Arthur Gessler, Michael~James McCarthy, Francesca Pellicciotti, Benjamin Stocker, and Dirk~Nikolaus Karger.
\newblock Global increase in the occurrence and impact of multiyear droughts.
\newblock \emph{Science}, 387\penalty0 (6731):\penalty0 278--284, 2025.

\bibitem[Dupont et~al.(2022)Dupont, Loya, Alizadeh, Goli{\'n}ski, Teh, and Doucet]{dupont_2022_COIN}
Emilien Dupont, Hrushikesh Loya, Milad Alizadeh, Adam Goli{\'n}ski, Yee~Whye Teh, and Arnaud Doucet.
\newblock Coin++: Neural compression across modalities, 2022.

\bibitem[Esser et~al.(2020)Esser, Rombach, and Ommer]{esser_2020_Taming}
Patrick Esser, Robin Rombach, and Bj{\"o}rn Ommer.
\newblock Taming transformers for high-resolution image synthesis, 2020.

\bibitem[Gong et~al.(2024{\natexlab{a}})Gong, Bai, Ye, Xu, Liu, Dai, Yang, and Ouyang]{gong2024cascast}
Junchao Gong, Lei Bai, Peng Ye, Wanghan Xu, Na Liu, Jianhua Dai, Xiaokang Yang, and Wanli Ouyang.
\newblock Cascast: Skillful high-resolution precipitation nowcasting via cascaded modelling.
\newblock In \emph{International Conference on Machine Learning}, pages 15809--15822. PMLR, 2024{\natexlab{a}}.

\bibitem[Gong et~al.(2024{\natexlab{b}})Gong, Tu, Yang, Fei, Chen, Zhang, Yang, Ouyang, and Bai]{gong2024postcast}
Junchao Gong, Siwei Tu, Weidong Yang, Ben Fei, Kun Chen, Wenlong Zhang, Xiaokang Yang, Wanli Ouyang, and Lei Bai.
\newblock Postcast: Generalizable postprocessing for precipitation nowcasting via unsupervised blurriness modeling.
\newblock \emph{arXiv preprint arXiv:2410.05805}, 2024{\natexlab{b}}.

\bibitem[Han et~al.(2024{\natexlab{a}})Han, Liu, Jiang, Yan, Zhang, Yuan, Peng, and Liu]{han2024infinity}
Jian Han, Jinlai Liu, Yi Jiang, Bin Yan, Yuqi Zhang, Zehuan Yuan, Bingyue Peng, and Xiaobing Liu.
\newblock Infinity: Scaling bitwise autoregressive modeling for high-resolution image synthesis.
\newblock \emph{arXiv preprint arXiv:2412.04431}, 2024{\natexlab{a}}.

\bibitem[Han et~al.(2024{\natexlab{b}})Han, Chen, Guo, Xu, and Bai]{han_2024_CRA5}
Tao Han, Zhenghao Chen, Song Guo, Wanghan Xu, and Lei Bai.
\newblock Cra5: Extreme compression of era5 for portable global climate and weather research via an efficient variational transformer, 2024{\natexlab{b}}.

\bibitem[Han et~al.(2024{\natexlab{c}})Han, Guo, Ling, Chen, Gong, Luo, Gu, Dai, Ouyang, and Bai]{han_2024_FengWuGHR}
Tao Han, Song Guo, Fenghua Ling, Kang Chen, Junchao Gong, Jingjia Luo, Junxia Gu, Kan Dai, Wanli Ouyang, and Lei Bai.
\newblock Fengwu-ghr: Learning the kilometer-scale medium-range global weather forecasting, 2024{\natexlab{c}}.

\bibitem[{Hansen-Estruch} et~al.(2025){Hansen-Estruch}, Yan, Chung, Zohar, Wang, Hou, Xu, Vishwanath, Vajda, and Chen]{hansen-estruch_2025_Learnings}
Philippe {Hansen-Estruch}, David Yan, Ching-Yao Chung, Orr Zohar, Jialiang Wang, Tingbo Hou, Tao Xu, Sriram Vishwanath, Peter Vajda, and Xinlei Chen.
\newblock Learnings from scaling visual tokenizers for reconstruction and generation, 2025.

\bibitem[He et~al.(2022)He, Yang, Peng, Ma, Qin, and Wang]{he_2022_ELIC}
Dailan He, Ziming Yang, Weikun Peng, Rui Ma, Hongwei Qin, and Yan Wang.
\newblock Elic: Efficient learned image compression with unevenly grouped space-channel contextual adaptive coding, 2022.

\bibitem[Hersbach et~al.(2020)Hersbach, Bell, Berrisford, Hirahara, Hor{\'a}nyi, Mu{\~n}oz-Sabater, Nicolas, Peubey, Radu, Schepers, Simmons, Soci, Abdalla, Abellan, Balsamo, Bechtold, Biavati, Bidlot, Bonavita, De~Chiara, Dahlgren, Dee, Diamantakis, Dragani, Flemming, Forbes, Fuentes, Geer, Haimberger, Healy, Hogan, H{\'o}lm, Janiskov{\'a}, Keeley, Laloyaux, Lopez, Lupu, Radnoti, De~Rosnay, Rozum, Vamborg, Villaume, and Th{\'e}paut]{hersbach_2020_ERA5}
Hans Hersbach, Bill Bell, Paul Berrisford, Shoji Hirahara, Andr{\'a}s Hor{\'a}nyi, Joaqu{\'i}n Mu{\~n}oz-Sabater, Julien Nicolas, Carole Peubey, Raluca Radu, Dinand Schepers, Adrian Simmons, Cornel Soci, Saleh Abdalla, Xavier Abellan, Gianpaolo Balsamo, Peter Bechtold, Gionata Biavati, Jean Bidlot, Massimo Bonavita, Giovanna De~Chiara, Per Dahlgren, Dick Dee, Michail Diamantakis, Rossana Dragani, Johannes Flemming, Richard Forbes, Manuel Fuentes, Alan Geer, Leo Haimberger, Sean Healy, Robin~J. Hogan, El{\'i}as H{\'o}lm, Marta Janiskov{\'a}, Sarah Keeley, Patrick Laloyaux, Philippe Lopez, Cristina Lupu, Gabor Radnoti, Patricia De~Rosnay, Iryna Rozum, Freja Vamborg, Sebastien Villaume, and Jean-No{\"e}l Th{\'e}paut.
\newblock The era5 global reanalysis.
\newblock \emph{Quarterly Journal of the Royal Meteorological Society}, 146\penalty0 (730):\penalty0 1999--2049, 2020.

\bibitem[Hua and {Chong-Yin}(2010)]{hua_2010_Further}
Tian Hua and Li {Chong-Yin}.
\newblock Further study of typhoon tracks and the low-frequency (30-60 days) wind-field pattern at 850 hpa.
\newblock \emph{Atmospheric and Oceanic Science Letters}, 3\penalty0 (6):\penalty0 319--324, 2010.

\bibitem[Huang and Hoefler(2022)]{huang_2022_Compressing}
Langwen Huang and Torsten Hoefler.
\newblock Compressing multidimensional weather and climate data into neural networks, 2022.

\bibitem[Inness et~al.(2019)Inness, Ades, {Agust{\'i}-Panareda}, Barr{\'e}, Benedictow, Blechschmidt, Dominguez, Engelen, Eskes, Flemming, Huijnen, Jones, Kipling, Massart, Parrington, Peuch, Razinger, Remy, Schulz, and Suttie]{inness_2019_CAMS}
Antje Inness, Melanie Ades, Anna {Agust{\'i}-Panareda}, J{\'e}r{\^o}me Barr{\'e}, Anna Benedictow, Anne-Marlene Blechschmidt, Juan~Jose Dominguez, Richard Engelen, Henk Eskes, Johannes Flemming, Vincent Huijnen, Luke Jones, Zak Kipling, Sebastien Massart, Mark Parrington, Vincent-Henri Peuch, Miha Razinger, Samuel Remy, Michael Schulz, and Martin Suttie.
\newblock The cams reanalysis of atmospheric composition.
\newblock \emph{Atmospheric Chemistry and Physics}, 19\penalty0 (6):\penalty0 3515--3556, 2019.

\bibitem[Kl{\"o}wer et~al.(2021)Kl{\"o}wer, Razinger, Dominguez, D{\"u}ben, and Palmer]{klöwer_2021_Compressing}
Milan Kl{\"o}wer, Miha Razinger, Juan~J. Dominguez, Peter~D. D{\"u}ben, and Tim~N. Palmer.
\newblock Compressing atmospheric data into its real information content.
\newblock \emph{Nature Computational Science}, 1\penalty0 (11):\penalty0 713--724, 2021.

\bibitem[Kuligowski and Barros(1998)]{kuligowski_1998_Experiments}
Robert~J. Kuligowski and Ana~P. Barros.
\newblock Experiments in short-term precipitation forecasting using artificial neural networks.
\newblock \emph{Monthly Weather Review}, 126\penalty0 (2):\penalty0 470--482, 1998.

\bibitem[Kurth et~al.(2023)Kurth, Subramanian, Harrington, Pathak, Mardani, Hall, Miele, Kashinath, and Anandkumar]{kurth2023fourcastnet}
Thorsten Kurth, Shashank Subramanian, Peter Harrington, Jaideep Pathak, Morteza Mardani, David Hall, Andrea Miele, Karthik Kashinath, and Anima Anandkumar.
\newblock Fourcastnet: Accelerating global high-resolution weather forecasting using adaptive fourier neural operators.
\newblock In \emph{Proceedings of the platform for advanced scientific computing conference}, pages 1--11, 2023.

\bibitem[Liang et~al.(2023)Liang, Zhao, Di, Li, Underwood, Gok, Tian, Deng, Calhoun, Tao, Chen, and Cappello]{liang_2023_SZ3}
Xin Liang, Kai Zhao, Sheng Di, Sihuan Li, Robert Underwood, Ali~M. Gok, Jiannan Tian, Junjing Deng, Jon~C. Calhoun, Dingwen Tao, Zizhong Chen, and Franck Cappello.
\newblock Sz3: A modular framework for composing prediction-based error-bounded lossy compressors.
\newblock \emph{IEEE Transactions on Big Data}, 9\penalty0 (2):\penalty0 485--498, 2023.

\bibitem[Liu et~al.(2022)Liu, Yan, Tong, Jiang, Li, Xia, Lou, Ren, and Fang]{liu_2022_Meshless}
Nian Liu, Zhongwei Yan, Xuan Tong, Jiang Jiang, Haochen Li, Jiangjiang Xia, Xiao Lou, Rui Ren, and Yi Fang.
\newblock Meshless surface wind speed field reconstruction based on machine learning.
\newblock \emph{Advances in Atmospheric Sciences}, 39\penalty0 (10):\penalty0 1721--1733, 2022.

\bibitem[Loshchilov and Hutter(2017)]{loshchilov2017decoupled}
Ilya Loshchilov and Frank Hutter.
\newblock Decoupled weight decay regularization.
\newblock \emph{arXiv preprint arXiv:1711.05101}, 2017.

\bibitem[Mentzer et~al.(2019)Mentzer, Agustsson, Tschannen, Timofte, and Van~Gool]{mentzer_2019_Practical}
Fabian Mentzer, Eirikur Agustsson, Michael Tschannen, Radu Timofte, and Luc Van~Gool.
\newblock Practical full resolution learned lossless image compression.
\newblock In \emph{2019 Ieee/Cvf Conference on Computer Vision and Pattern Recognition (Cvpr)}, pages 10621--10630, Long Beach, CA, USA, 2019. IEEE.

\bibitem[Mirowski et~al.(2024)Mirowski, {Warde-Farley}, Rosca, Grimes, Hasson, Kim, Rey, Osindero, Ravuri, and Mohamed]{mirowski_2024_Neural}
Piotr Mirowski, David {Warde-Farley}, Mihaela Rosca, Matthew~Koichi Grimes, Yana Hasson, Hyunjik Kim, M{\'e}lanie Rey, Simon Osindero, Suman Ravuri, and Shakir Mohamed.
\newblock Neural compression of atmospheric states, 2024.

\bibitem[Moore and Dixon(2015)]{moore_2015_Patterns}
Todd~W. Moore and Richard~W. Dixon.
\newblock Patterns in 500 hpa geopotential height associated with temporal clusters of tropical cyclone tornadoes: 500 hpa geopotential height patterns and tropical cyclone tornadoes.
\newblock \emph{Meteorological Applications}, 22\penalty0 (3):\penalty0 314--322, 2015.

\bibitem[Patz et~al.(2005)Patz, {Campbell-Lendrum}, Holloway, and Foley]{patz_2005_Impact}
Jonathan~A. Patz, Diarmid {Campbell-Lendrum}, Tracey Holloway, and Jonathan~A. Foley.
\newblock Impact of regional climate change on human health.
\newblock \emph{Nature}, 438\penalty0 (7066):\penalty0 310--317, 2005.

\bibitem[Price et~al.(2023)Price, {Sanchez-Gonzalez}, Alet, Andersson, {El-Kadi}, Masters, Ewalds, Stott, Mohamed, Battaglia, Lam, and Willson]{price_2023_GenCast}
Ilan Price, Alvaro {Sanchez-Gonzalez}, Ferran Alet, Tom~R. Andersson, Andrew {El-Kadi}, Dominic Masters, Timo Ewalds, Jacklynn Stott, Shakir Mohamed, Peter Battaglia, Remi Lam, and Matthew Willson.
\newblock Gencast: Diffusion-based ensemble forecasting for medium-range weather, 2023.

\bibitem[Rasp et~al.(2020)Rasp, Dueben, Scher, Weyn, Mouatadid, and Thuerey]{rasp_2020_WeatherBench}
Stephan Rasp, Peter~D. Dueben, Sebastian Scher, Jonathan~A. Weyn, Soukayna Mouatadid, and Nils Thuerey.
\newblock Weatherbench: A benchmark data set for data-driven weather forecasting.
\newblock \emph{Journal of Advances in Modeling Earth Systems}, 12\penalty0 (11):\penalty0 e2020MS002203, 2020.

\bibitem[Ravuri et~al.(2021)Ravuri, Lenc, Willson, Kangin, Lam, Mirowski, Fitzsimons, Athanassiadou, Kashem, Madge, Prudden, Mandhane, Clark, Brock, Simonyan, Hadsell, Robinson, Clancy, Arribas, and Mohamed]{ravuri_2021_Skilful}
Suman Ravuri, Karel Lenc, Matthew Willson, Dmitry Kangin, Remi Lam, Piotr Mirowski, Megan Fitzsimons, Maria Athanassiadou, Sheleem Kashem, Sam Madge, Rachel Prudden, Amol Mandhane, Aidan Clark, Andrew Brock, Karen Simonyan, Raia Hadsell, Niall Robinson, Ellen Clancy, Alberto Arribas, and Shakir Mohamed.
\newblock Skilful precipitation nowcasting using deep generative models of radar.
\newblock \emph{Nature}, 597\penalty0 (7878):\penalty0 672--677, 2021.

\bibitem[Tian et~al.(2015)Tian, Zheng, Zhang, Zhang, Mao, Sun, and Zhao]{tian_2015_Statistical}
Fuyou Tian, Yongguang Zheng, Tao Zhang, Xiaoling Zhang, Dongyan Mao, Jianhua Sun, and Sixiong Zhao.
\newblock Statistical characteristics of environmental parameters for warm season short-duration heavy rainfall over central and eastern china.
\newblock \emph{Journal of Meteorological Research}, 29\penalty0 (3):\penalty0 370--384, 2015.

\bibitem[Van Den~Oord et~al.(2017)Van Den~Oord, Vinyals, et~al.]{van2017neural}
Aaron Van Den~Oord, Oriol Vinyals, et~al.
\newblock Neural discrete representation learning.
\newblock \emph{Advances in neural information processing systems}, 30, 2017.

\bibitem[Wild et~al.(2025)Wild, Huey, Pianka, {Clusella-Trullas}, Gilbert, Miles, and Kearney]{wild_2025_Climate}
Kristoffer~H. Wild, Raymond~B. Huey, Eric~R. Pianka, Susana {Clusella-Trullas}, Anthony~L. Gilbert, Donald~B. Miles, and Michael~R. Kearney.
\newblock Climate change and the cost-of-living squeeze in desert lizards.
\newblock \emph{Science}, 387\penalty0 (6731):\penalty0 303--309, 2025.

\bibitem[Xie et~al.(2021)Xie, Cheng, and Chen]{xie_2021_Enhanced}
Yueqi Xie, Ka~Leong Cheng, and Qifeng Chen.
\newblock Enhanced invertible encoding for learned image compression, 2021.

\bibitem[Xiong et~al.(2024)Xiong, Wang, Zhang, Stewart, Hanna, Borth, Papoutsis, Saux, {Camps-Valls}, and Zhu]{xiong_2024_Neural}
Zhitong Xiong, Yi Wang, Fahong Zhang, Adam~J. Stewart, Jo{\"e}lle Hanna, Damian Borth, Ioannis Papoutsis, Bertrand~Le Saux, Gustau {Camps-Valls}, and Xiao~Xiang Zhu.
\newblock Neural plasticity-inspired multimodal foundation model for earth observation, 2024.

\bibitem[Xu et~al.(2024)Xu, Chen, Han, Chen, Ouyang, and Bai]{xu2024extremecast}
Wanghan Xu, Kang Chen, Tao Han, Hao Chen, Wanli Ouyang, and Lei Bai.
\newblock Extremecast: Boosting extreme value prediction for global weather forecast.
\newblock \emph{arXiv preprint arXiv:2402.01295}, 2024.

\bibitem[Yang et~al.(2023)Yang, Liu, Hu, Liu, Xie, and Zhao]{yang_2023_Predictor}
Dangfu Yang, Shengjun Liu, Yamin Hu, Xinru Liu, Jiehong Xie, and Liang Zhao.
\newblock Predictor selection for cnn-based statistical downscaling of monthly precipitation.
\newblock \emph{Advances in Atmospheric Sciences}, 40\penalty0 (6):\penalty0 1117--1131, 2023.

\bibitem[Zhang et~al.(2023{\natexlab{a}})Zhang, Cao, Chu, Zhao, Zhao, and Lee]{zhang2023increased}
Keer Zhang, Chang Cao, Haoran Chu, Lei Zhao, Jiayu Zhao, and Xuhui Lee.
\newblock Increased heat risk in wet climate induced by urban humid heat.
\newblock \emph{Nature}, 617\penalty0 (7962):\penalty0 738--742, 2023{\natexlab{a}}.

\bibitem[Zhang et~al.(2023{\natexlab{b}})Zhang, Long, Chen, Xing, Jin, Jordan, and Wang]{zhang_2023_Skilful}
Yuchen Zhang, Mingsheng Long, Kaiyuan Chen, Lanxiang Xing, Ronghua Jin, Michael~I. Jordan, and Jianmin Wang.
\newblock Skilful nowcasting of extreme precipitation with nowcastnet.
\newblock \emph{Nature}, 619\penalty0 (7970):\penalty0 526--532, 2023{\natexlab{b}}.

\bibitem[Zhao et~al.(2024)Zhao, Xiong, and Kr{\"a}henb{\"u}hl]{zhao_2024_Image}
Yue Zhao, Yuanjun Xiong, and Philipp Kr{\"a}henb{\"u}hl.
\newblock Image and video tokenization with binary spherical quantization, 2024.

\end{thebibliography}
}

\ifarxiv 
\clearpage 
\appendix 
\maketitlesupplementary
\renewcommand{\thefigure}{S\arabic{figure}}  
\setcounter{figure}{0}  

\section{Visualization of Reconstructed Results}
\label{sec:appendix_section}

We visualize several variables from the ERA5 dataset and its compressed representation (ERA5-latent) in Figures S1 to S6, where the Mean Absolute Error maps highlight their differences across regions. These examples demonstrate that the ERA5-latent dataset effectively captures representative features, preserving detailed information in atmospheric data.

\begin{figure*}[!h]
    \centering
    \includegraphics[width=1.0\textwidth]{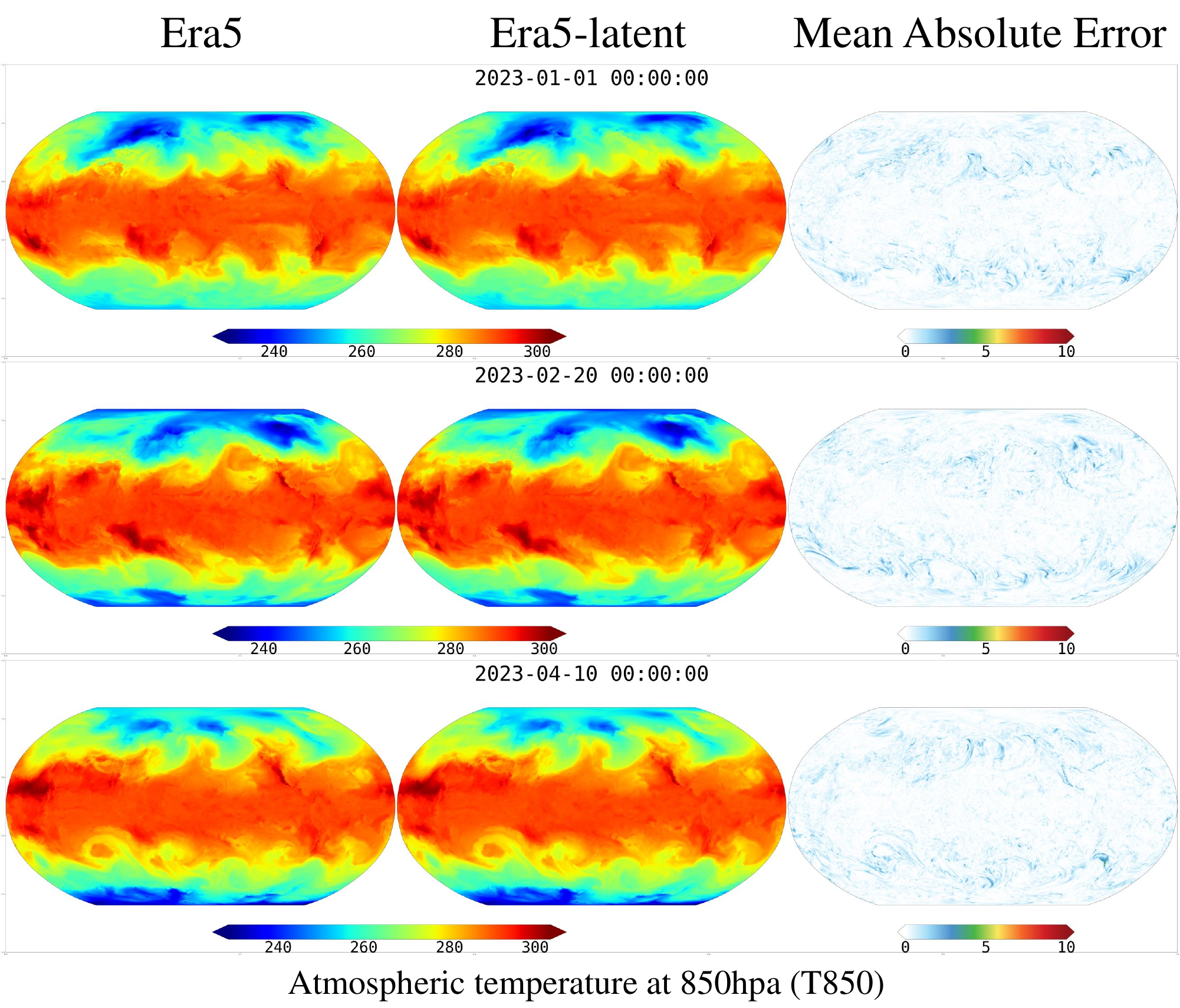}
    \caption{Visualization samples of T850 on the ERA5 and the compressed ERA5-latent. From the left to the right column: ERA5, ERA5-latent, and their absolute error map.}
    \label{fig:S1}
\end{figure*}

\begin{figure*}[!ht]
    \centering
    \includegraphics[width=1.0\textwidth]{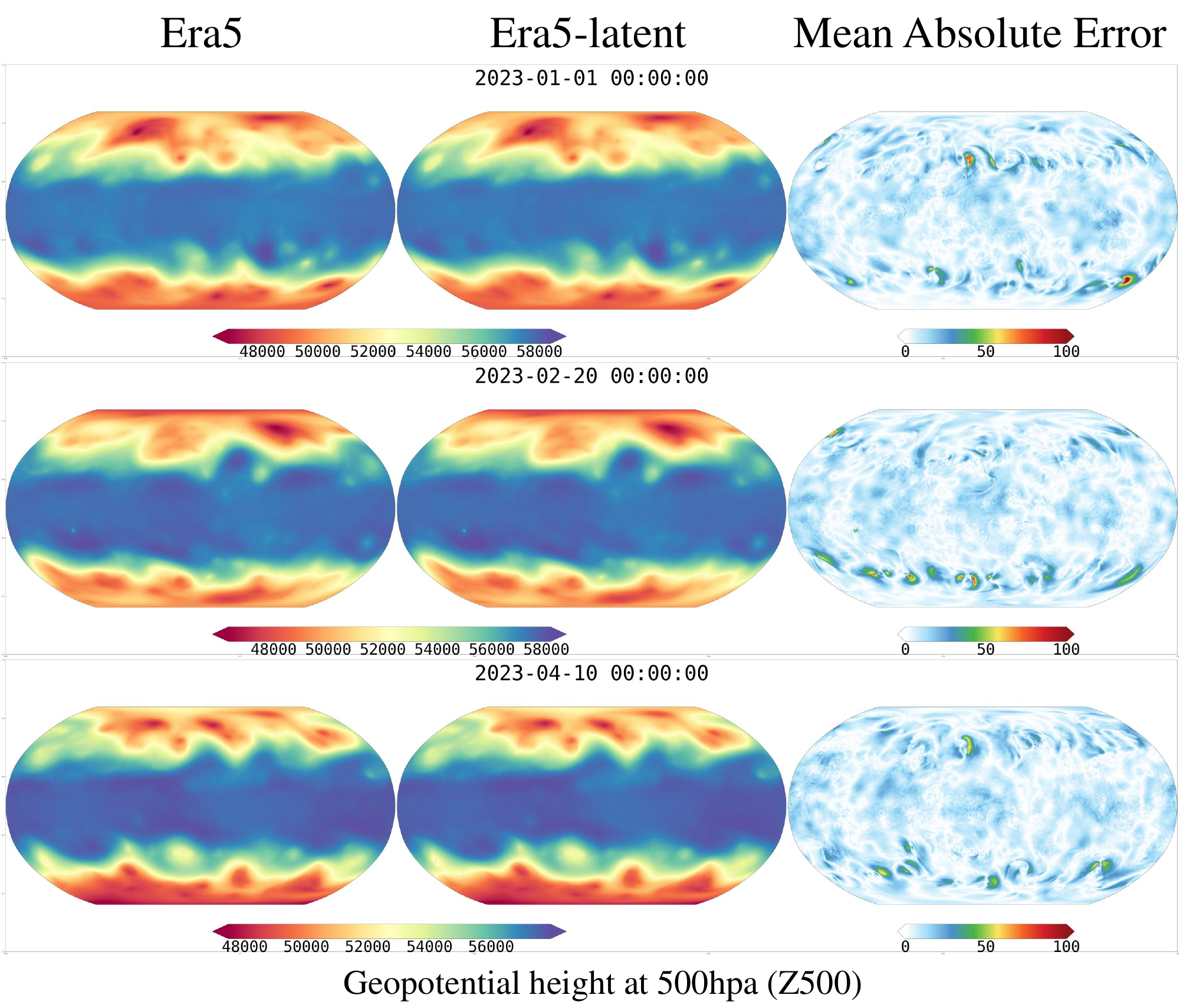}
    \caption{Visualization samples of Z500 on the ERA5 and the compressed ERA5-latent. From the left to the right column: ERA5, ERA5-latent, and their absolute error map.}
    \label{fig:S2}
\end{figure*}

\begin{figure*}[!ht]
    \centering
    \includegraphics[width=1.0\textwidth]{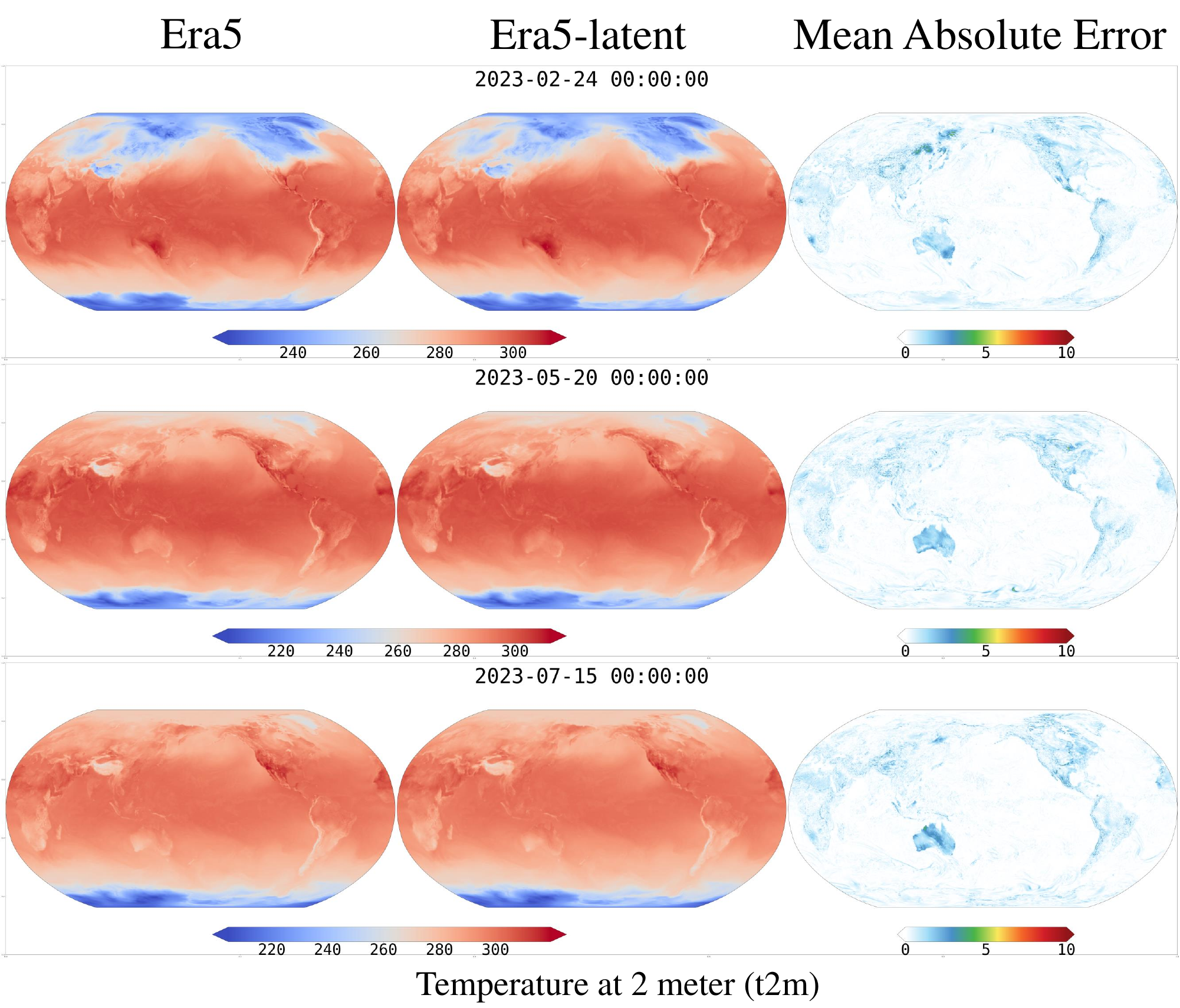}
    \caption{Visualization samples of t2m on the ERA5 and the compressed ERA5-latent. From the left to the right column: ERA5, ERA5-latent, and their absolute error map.}
    \label{fig:S3}
\end{figure*}

\begin{figure*}[!ht]
    \centering
    \includegraphics[width=1.0\textwidth]{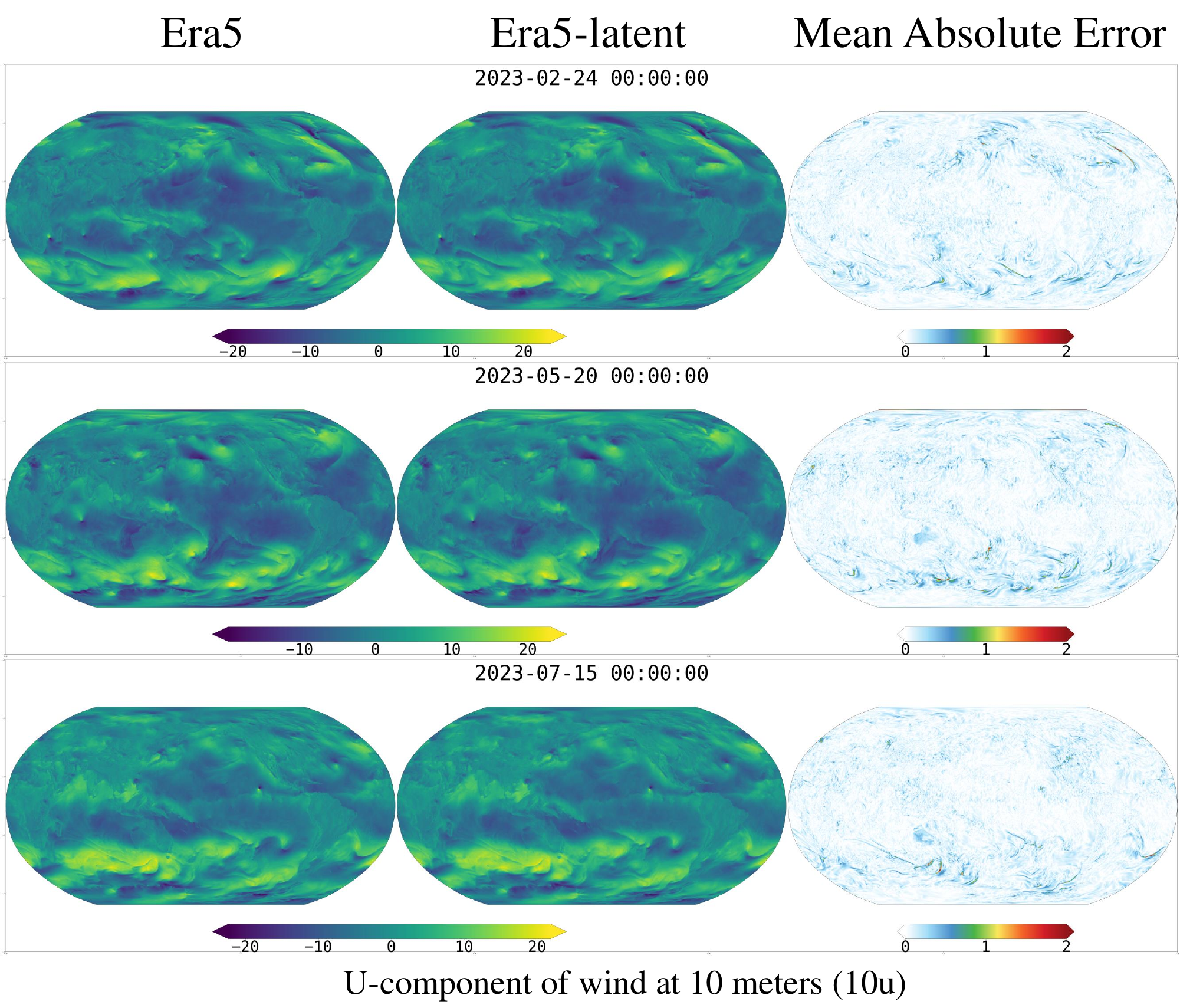}
    \caption{Visualization samples of 10u on the ERA5 and the compressed ERA5-latent. From the left to the right column: ERA5, ERA5-latent, and their absolute error map.}
    \label{fig:S4}
\end{figure*}

\begin{figure*}[!ht]
    \centering
    \includegraphics[width=1.0\textwidth]{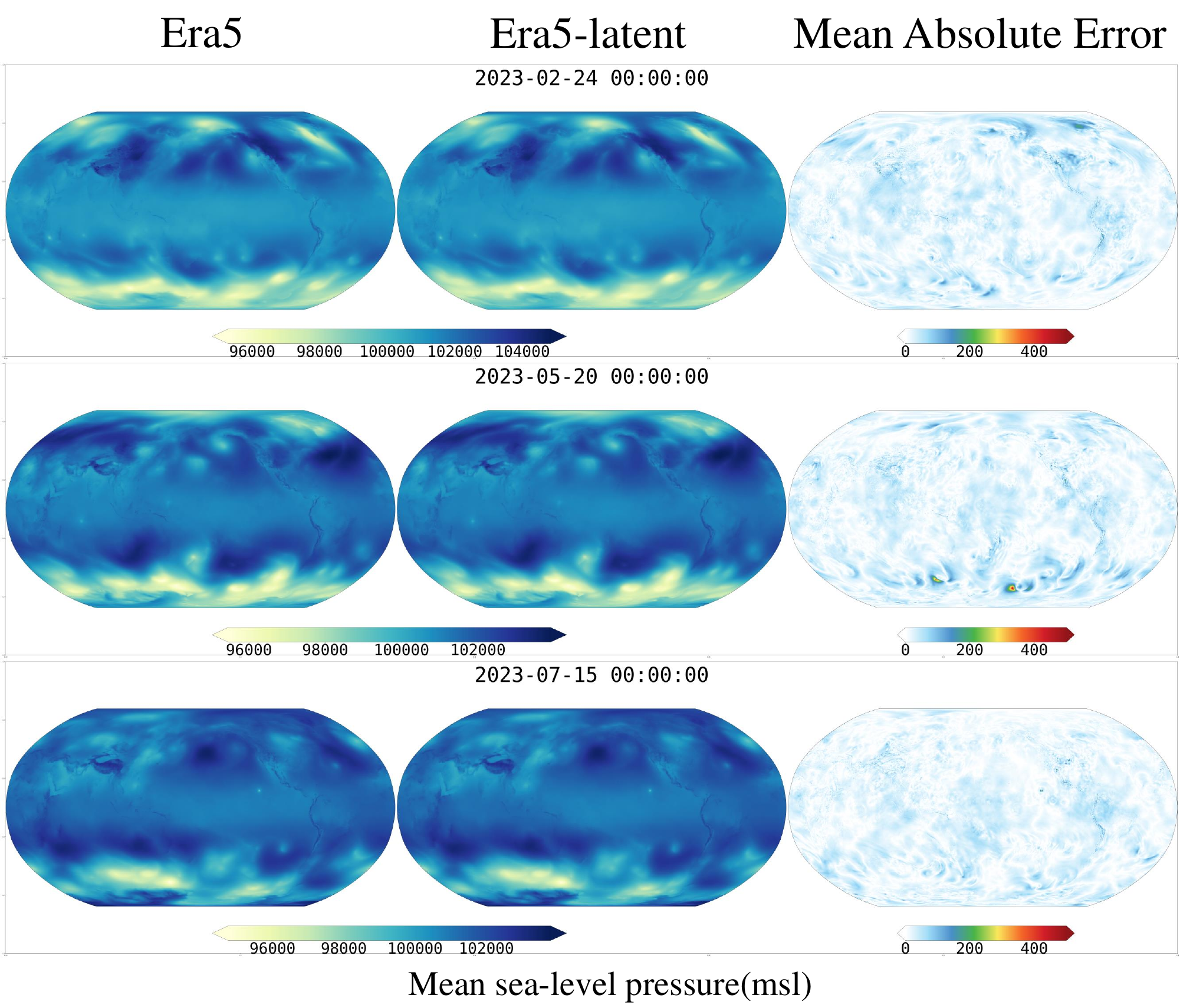}
    \caption{Visualization samples of msl on the ERA5 and the compressed ERA5-latent. From the left to the right column: ERA5, ERA5-latent, and their absolute error map.}
    \label{fig:S5}
\end{figure*}

\begin{figure*}[!ht]
    \centering
    \includegraphics[width=1.0\textwidth]{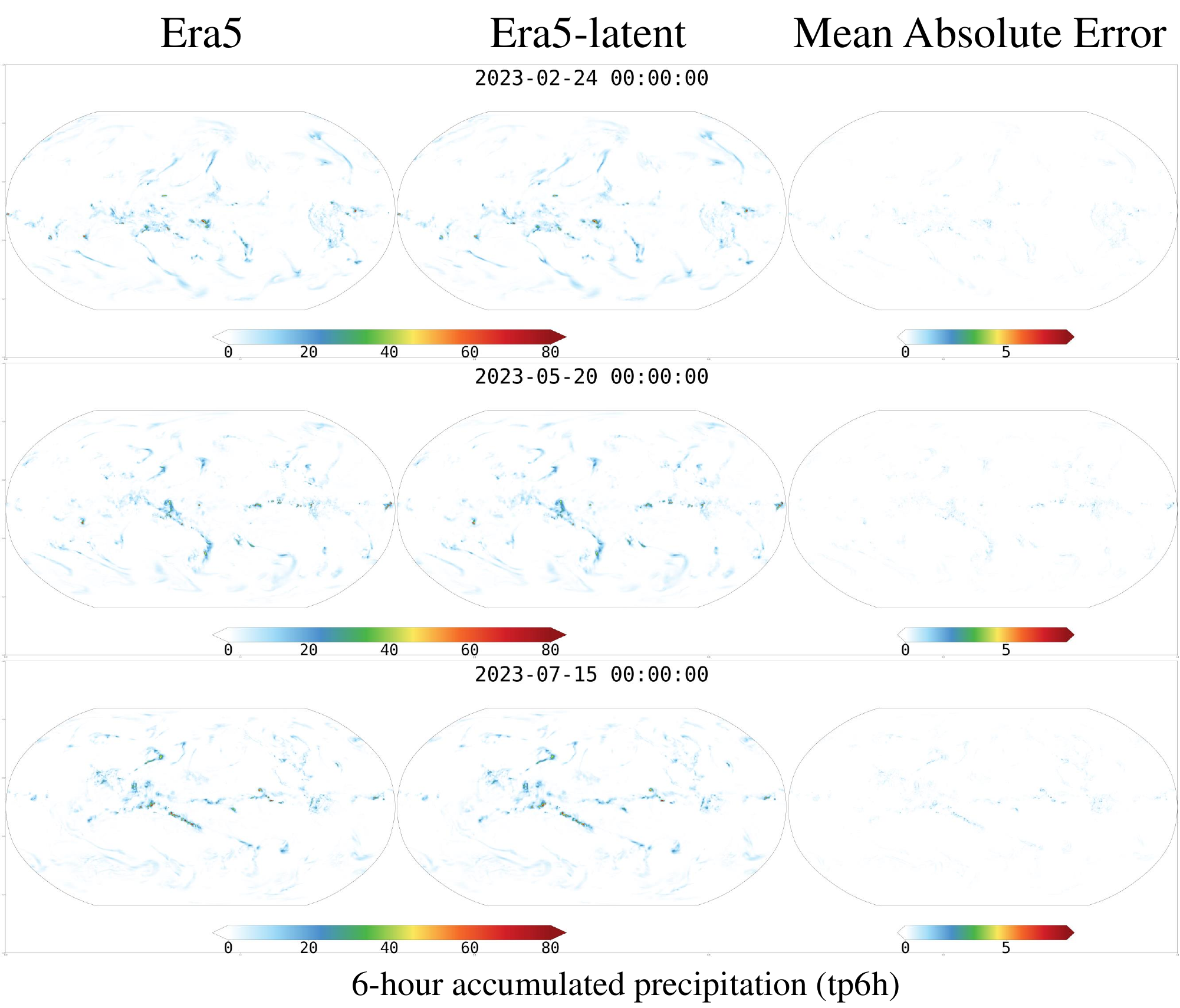}
    \caption{Visualization samples of tp6h on the ERA5 and the compressed ERA5-latent. From the left to the right column: ERA5, ERA5-latent, and their absolute error map.}
    \label{fig:S6}
\end{figure*} 
\fi

\end{document}